\documentclass{article}

\usepackage[final]{corl_2018}

\usepackage{amsmath, amsfonts,bm}
\usepackage{graphicx}
\usepackage{wrapfig}
\usepackage{verbatim}
\usepackage{float}

\usepackage{color, colortbl}
\definecolor{LightCyan}{rgb}{0.88,1,1}
\definecolor{Gray}{gray}{0.9}

\def\Real{\mathbb{R}}
\renewcommand\Pr{{\operatorname{Pr}}}

\newcommand\deq{\overset{\mathrm{def}}{=\joinrel=}}
\def\ov{{\bf o}}

\title{Benchmarking Reinforcement Learning Algorithms on Real-World Robots}

\author{
A. Rupam Mahmood \\
\texttt{rupam@kindred.ai} \\
 \And
 Dmytro Korenkevych \\
\texttt{dmytro.korenkevych@kindred.ai } \\
 \AND
 Gautham Vasan \\
\texttt{gautham.vasan@kindred.ai} \\ 
\And
 William Ma \\
\texttt{william.ma@kindred.ai} \\
 \And 
 James Bergstra \\
 \texttt{james@kindred.ai} \\
}

\begin{document}
\maketitle


\begin{abstract}

Through many recent successes in simulation, model-free reinforcement learning has emerged as a promising approach to solving continuous control robotic tasks.  
The research community is now able to reproduce, analyze and build quickly on these results due to open source implementations of learning algorithms and simulated benchmark tasks. To carry forward these successes to real-world applications, it is crucial to withhold utilizing the unique advantages of simulations that do not transfer to the real world and experiment directly with physical robots. 
However, reinforcement learning research with physical robots faces substantial resistance due to the lack of benchmark tasks and supporting source code. 
In this work, we introduce several reinforcement learning tasks with multiple commercially available robots that present varying levels of learning difficulty, setup, and repeatability.
On these tasks, we test the learning performance of off-the-shelf implementations of four reinforcement learning algorithms and analyze sensitivity to their hyper-parameters to determine their readiness for applications in various real-world tasks. 
Our results show that with a careful setup of the task interface and computations, some of these implementations can be readily applicable to physical robots. 
We find that state-of-the-art learning algorithms are highly sensitive to their hyper-parameters and their relative ordering does not transfer across tasks, indicating the necessity of re-tuning them for each task for best performance.
On the other hand, the best hyper-parameter configuration from one task may often result in effective learning on held-out tasks even with different robots, providing a reasonable default.
We make the benchmark tasks publicly available to enhance reproducibility in real-world reinforcement learning\footnote{Source code for all tasks available at \url{https://github.com/kindredresearch/SenseAct}}.

\end{abstract}



\section{Introduction}

In recent years, policy learning algorithms such as trust region policy optimization (TRPO, Schulman et al.\ 2015), proximal policy optimization (PPO, Schulman et al.\ 2016), and deep deterministic policy gradient (DDPG, Lillicrap et al.\ 2015) methods have gained popularity due to their success in various simulated robotic tasks (Duan et al.\ 2016).
A large body of works has been built on these algorithms to address different challenges in reinforcement learning including policy learning (Haarnoja et al. 2017), hierarchical learning (Klissarov et al.\ 2017), transfer learning (Wulfmeier, Posner \& Abbeel 2017), and emergence of complex behavior (Heess et al.\ 2017).
Deep learning software such as Theano and Tensorflow as well as the availability of source code of learning algorithms (e.g., Duan et al.\ 2016, Dhariwal et al.\ 2017) and benchmark simulated environments (e.g., Brockman et al.\ 2016, Machado et al.\ 2018, Tassa et al.\ 2018) contributed to this advancement.

It is natural to expect that successes in simulations would inspire similar engagement within the reinforcement learning community toward policy learning with physical robots.
But this engagement so far has been limited.
Some notable works show success when the learning algorithms are supported with one or more of
(a) sufficient state information or auxiliary task-specific steps and knowledge (e.g.\ Levine et~al.~2016, Riedmiller et al.\ 2018),
(b) preparation in simulation (e.g.\ Rusu et al.\ 2017)
(c) collaborative learning (e.g.\ Yahya et al.\ 2017),
and (d) learning from demonstrations (e.g.\ Hester et al.\ 2017).
However, reinforcement learning research with real-world robots is yet to fully embrace and engage the purest and simplest form of the reinforcement learning problem statement---an agent maximizing its rewards by learning from its first-hand experience of the world.
This lack of engagement indicates the difficulties in carrying forward the successes and enthusiasm found in simulation-based works to the real world.
Due to the lack of benchmark tasks, it is hard to analyze these difficulties and address them as a community.

Mahmood et al.\ (2018) recently brought to attention some of the difficulties of real-world robot learning and showed that learning performance can be highly sensitive to different elements of the task setup such as the action space, the action cycle time defined by the time between two subsequent actions, and system delays (also see Riedmiller 2012).
Therefore, reproducing and utilizing existing results can be hard when the details of these task setup elements are omitted.
Moreover, without careful task setups, learning with physical robots can be insurmountably difficult.

To study and alleviate these difficulties, we introduce six reinforcement learning tasks based on three commercially available robots.
Most of these tasks require no additional hardware installation apart from the basic robot setup.
On these tasks, we compare and benchmark four reinforcement learning algorithms for continuous control: TRPO, PPO, DDPG, and Soft Q-learning (Haarnoja et al.\ 2017).
The main contributions of this work are 1) introducing benchmark tasks for physical robots to share across the community, 2) setting up the tasks to be conducive to learning, and 3) providing the first extensive empirical study of multiple policy learning algorithms on multiple physical robots.

\section{Robots}

\begin{figure}  
\center
\includegraphics[scale=.54]{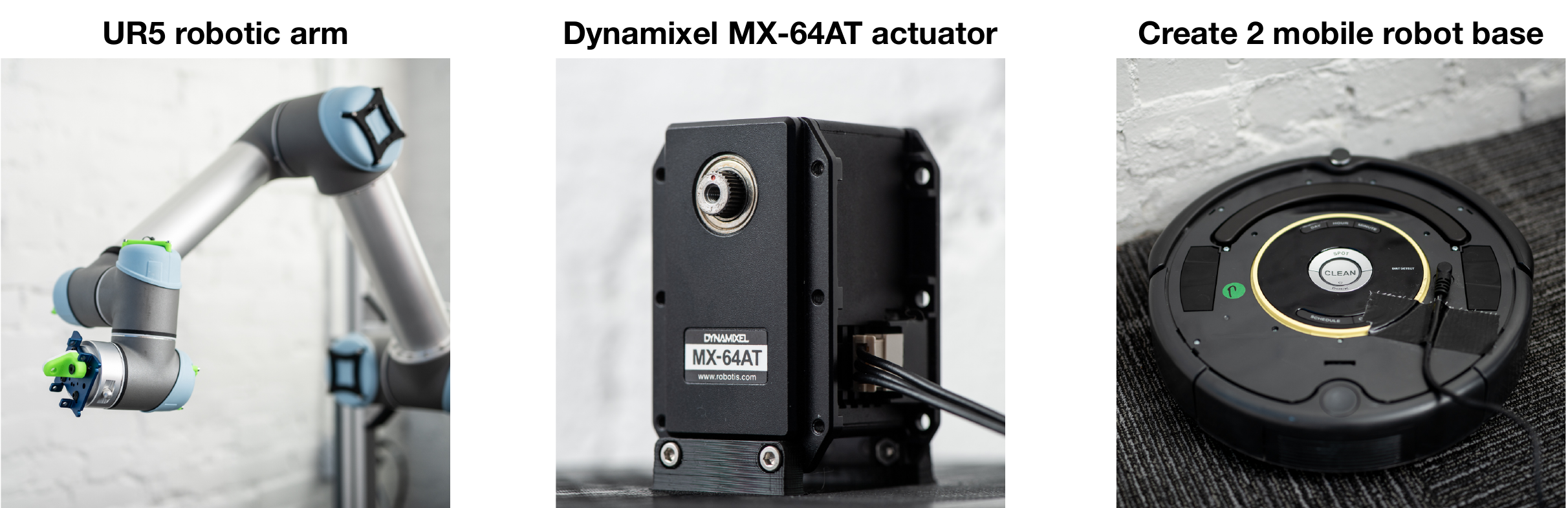}
\caption{
    {\bf The robots used in this work:} (\emph{left}) Universal Robotics UR5 collaborative arms, (\emph{middle}) Robotis MX-64AT Dynamixel actuators, and (\emph{right}) iRobot Create2 mobile robot bases.
}
\label{fig:robots}
\end{figure}

We use three commercially available robots (see Figure \ref{fig:robots}) as a basis for defining learning tasks. 

\textbf{UR5:} The UR5, shown in Figure \ref{fig:robots} (\emph{left}) is a collaborative industrial arm with six joints produced by Universal Robots.
The sensory packets from UR5 include angles, velocities, target accelerations, and currents for each joint.
The control interface offers low-level position and velocity control commands.
We use UR5 to develop two tasks called \emph{UR-Reacher-2} and \emph{UR-Reacher-6} based on the tasks developed by Mahmood et al.\ (2018).

\textbf{Dynamixel MX-64AT:} The Dynamixel (DXL) series of programmable Direct-Current actuators, manufactured by Robotis,
are popular for custom robots ranging from robot arms to humanoids.
We use single DXL MX-64AT actuators, shown in Figure \ref{fig:robots} (\emph{middle}), which complies with high torque and load requirements.
The MX series actuators are controlled by digital packets via a half duplex asynchronous serial communication protocol, that is, we can read and write to the motor but not simultaneously.
The protocol allows a control computer to send position, velocity or current control commands (referred to as torque control in the manual) to the actuator as well as poll sensorimotor information including position, velocity, temperature, current and load.
We develop two tasks based on this actuator, which we call \emph{DXL-Reacher} and \emph{DXL-Tracker}.

\textbf{Create 2:} The Create 2, shown in Figure \ref{fig:robots} (\emph{right}), is a hobbyist version of iRobot's Roomba vacuum robot.
The Create 2 has two actuated wheels and many sensors including six front-facing infrared wall sensors, charging sensor, one omni-directional and two directional infrared sensors for docking, two physical bump sensors, and a directed distance sensor for the forward direction.
The software interface allows the control computer to access sensory packets in a streaming mode as well as send to the robot target speeds (mm/s) for its two wheels.
We develop two tasks with it, called \emph{Create-Mover} and \emph{Create-Docker}.

Appendix A.1 contains additional details of the hardware setups.

\section{Tasks}

In a reinforcement learning (RL) task (Sutton \& Barto 1998), an agent interacts with its environment at discrete time steps, where at each step $t$, the environment provides the agent its state information $S_t \in \cal{S}$ and a scalar reward signal $R_{t} \in \Real$.
The agent uses a stochastic policy $\pi$ with a probability distribution $\pi(a|s) \deq \Pr\left\{ A_t=a | S_t=s \right\}$ to select an action $A_t \in \cal{A}$.
In response, the environment transitions to a new state $S_{t+1}$ and produces a new reward $R_{t+1}$ at the next time step $t+1$ using a transition probability distribution: $p(s', r|s, a) \deq \Pr\left\{ S_{t+1} = s', R_{t+1}=r | S_t = s, A_t = a \right\}$.
The goal of the agent is typically to find a policy that maximizes the expected return defined as the future  accumulated rewards $G_t \deq \sum_{k=t}^\infty \gamma^{k-t} R_{k+1}$, where $\gamma\in[0,1]$ is a discount factor.
In practice, the agent observes the environment's state partially through a real-valued observation vector $\ov_t$ instead of receiving the state information fully.

\textbf{UR-Reacher-2:}
We use the Reacher task with UR5 developed by Mahmood et al.\ (2018), which is designed analogously to OpenAI-Gym Reacher (Brockman et al.\ 2016). 
We modify the reward function and call the task \emph{UR-Reacher-2}.
In Gym Reacher, the agent's objective is to reach arbitrary target positions by exercising low-level control over a two-joint robotic arm.
In UR-Reacher-2, we actuate the second and the third joints from the base by sending angular speeds between $[-0.3, +0.3]$ rad/s.
The observation vector consists of joint angles, joint velocities, the previous action, and the vector difference between the target and the fingertip coordinates.
The reward function is defined as: $R_{t} = -d_{t} + \exp(-100  d_{t}^2)$, where $d_t$ is the Euclidean distance between the target and the fingertip positions.
The second term of the reward function, which we call the \emph{precision reward}, incentivizes the algorithm to learn to get to the target with a high precision.
We defined episodes to be 4 seconds long to allow adequate exploration.
At each episode, the target position is generated randomly within a $0.7m\times 0.5m$ boundary, while the arm always starts from the middle of the boundary.

\textbf{UR-Reacher-6:}
The second task with UR5, which we call \emph{UR-Reacher-6}, is analogous to UR-Reacher-2 with the exceptions that all six joints are actuated and the target is drawn from a $0.7m\times 0.5m\times 0.4m$ 3D space.
The higher dimensionality of the action and observation spaces and physical limitations of reaching locations from various configurations of the arm joints in 3D space result in a much more complex policy space and substantially increase the learning problem difficulty.

\textbf{DXL-Reacher:}
We design a Reacher task similar to UR-Reacher-2 with current control of the DXL actuator, which we call \emph{DXL-Reacher}. 
The action space is one-dimensional current control signals between $[-100, 100]$ mA, making the task simpler than UR-Reacher-2.
The reward function is defined as: $R_{t} = -d_{t}$.
The observation vector includes the actuator position (in radians), moving speed, target position, and the previous action. 
Each episode is 2 seconds long to allow adequate time for reaching distant target positions. 
At each episode, the target position is chosen randomly within a certain boundary of angular positions, and the actuator starts from the center of it. 

\textbf{DXL-Tracker:}
We develop a second task using the DXL actuator, which we call \emph{DXL-Tracker}.
The objective here is to precisely track a moving target position with current control signals between $[-50, 50]$ mA.
The observation vector includes the actuator position (in radians), moving speed, current target position, target position from $50$ milliseconds in the past and the previous action. 
The reward function is same as that of DXL-Reacher.
Each episode is 4 seconds long to allow adequate time to catch up with the target and subsequently track it. 
At each episode, the starting position of the target is chosen uniformly randomly from a certain range, and the actuator starts from the center of that range. 
In addition, we also randomly choose the direction of the moving target.
The speed of the target is set in such a way that the target always arrives at a certain fixed position at the end of the episode. 
Thus, the speed is different for different episodes.

\textbf{Create-Mover:}
We develop a task with Create 2 where the agent needs to move the robot forward as fast as possible within an enclosed arena. 
We call it \emph{Create-Mover}.
A $3ft\times 2.5ft$ arena is built using white shelving boards for the walls and a white hardboard for the floor.
The action space is $[-150mm/s, 150mm/s]^2$ for actuating the two wheels with speed control.
The observation vector is composed of 6 wall-sensors values and the previous action.  
For the wall sensors, we always take the latest values received within the action cycle and use Equation 1 by (Benet et al.\ 2002) to convert the incoming signals to approximate distances.
The reward function is the summation of the directed distance values over 10 most recent sensory packets.  
An episode is 90 seconds long but ends earlier if the agent triggers one of its bump sensors.
When an episode terminates, the position of the robot is reset by moving backward to avoid bumping into the wall immediately.
We use two Create 2 robots and two identical arenas for our experiments. 
Among the two robots, one of them has two faulty wall sensors always receiving value zero, with four other sensors oriented symmetrically.
To make comparisons fair, each algorithm was run using both robots the same number of times.

\textbf{Create-Docker:} 
In this task the objective is to dock to a charging station attached to the middle of one of the wider walls of the Create-Mover arena. The reward function is a large positive number for successful docking with penalty for bumping and encouragement for moving forward and facing the charging station perpendicularly. More details of the task is provided in Appendix A.2.

All of these tasks are implemented following the computational model for real-time reinforcement learning tasks described by Mahmood et al.\ (2018).
We improve on that model by running environment and agent computations on two different processes, which we found to reduce execution delays compared to the use of Python threads.
The action cycle time is 150ms for Create-Mover, 45ms for Create-Docker and 40ms for the rest of the tasks.
The robot read-write cycle time is set to 8ms for UR5 tasks, 10ms for DXL tasks and 15ms for Create tasks.
The reward is scaled by the action cycle time in all cases. 
The action space is normalized between -1 and +1 for each dimension.
For the Create tasks, all the observations are also normalized between -1 and +1.

\section{Reinforcement learning algorithms}

We select four continuous control policy learning algorithms. 
TRPO, PPO and DDPG are among the most popular of their kind. 
On the other hand, Soft-Q is a new algorithm with promising results. 
We use the OpenAI Baselines implementations for TRPO, and PPO, Rllab implementation for DDPG, and the implementation of Soft Q-learning by the original authors.

\textbf{Trust region policy optimization (TRPO):}
TRPO (Schulman et al.\ 2015) is a policy optimization algorithm that constrain the change in the policy at each learning update.
The policy is optimized by iteratively solving the following constrained optimization problem:
\begin{align*}
\underset{\theta}{\text{maximize}}& 
\quad \mathbb{E}_{s, a \sim \pi_{\theta_\text{old}}}\left[r_\theta(a|s)A_{\theta_\text{old}}(s, a) \right], ~~~\text{s.t.}& \quad \mathbb{E}_{s, a \sim \pi_{\theta_\text{old}}}[D_\text{KL}(\pi_\theta(\cdot|s) \Vert \pi_{\theta_{\text{old}}}(\cdot|s))] \le \delta,  
\end{align*}
where $A_{\theta_\text{old}}$ is the advantage function, $r_\theta(a|s) = \frac{\pi_\theta(a | s)}{\pi_{\theta_\text{old}}(a | s)}$ is the ratio of a target policy probability to the policy probability used to generate data, $D_\text{KL}$ is a Kullback-Leibler divergence, and $\delta$ is a ``step-size'' parameter.
TRPO uses the conjugate-gradient algorithm to solve the above problem.

\textbf{Proximal policy optimization (PPO):}
PPO (Schulman et al.\ 2016) attempts to control the policy change during learning updates by replacing the KL-divergence constraint of TPRO in the optimization problem with a penalty term realized by a clipping in the objective function:
\begin{align*}
L^{CLIP}_\theta = \mathbb{E}_{s, a \sim \pi_{\theta_\text{old}}}\left[\min(r_\theta(a|s) A_{\theta_\text{old}}(a, s), \text{clip}(r_\theta(a|s), 1 - \varepsilon, 1 + \varepsilon) A_{\theta_\text{old}}(s, a)) \right], \\[0.5ex]
\end{align*}
where $A_{\theta_\text{old}}$ is the advantage function, and $\varepsilon$ is a parameter, usually on the order of 0.1.
The optimization is done by running several epochs of stochastic gradient ascent at each update.

\textbf{Soft Q-learning (Soft-Q):}
Soft-Q (Haarnoja et al. 2017) defines a policy as an energy based probability distribution: $\pi(a | s) \propto \exp( - \mathcal{E}(a, s))$, where the energy function $\mathcal{E}$ corresponds to a ``soft'' action-value function, obtained by optimizing the maximum entropy objective.
The soft action-value function is represented by deep neural networks, and therefore the policy energy model can represent complex multi-modal behaviors.
This model provides natural exploration mechanism without the need to introduce artificial sources of exploration noise, such as additive Gaussian noise.

\textbf{Deep deterministic policy gradient (DDPG):}
DDPG (Lillicrap et al.\ 2015) learns a deterministic policy that maximizes the estimated action-value function by following a policy gradient:
\begin{align*}
\nabla_{\theta} L_\theta = \frac{1}{N} \sum_i \nabla_a Q_\phi(s_i, a)|_{a = \mu_\theta(s_i)}\nabla_\theta \mu_\theta(s_i)
\end{align*}
where $N$ is the batch size.
In the Rllab implementation, the exploration of DDPG is addressed by augmenting the policy output $\mu_\theta(s)$ with additive noise from an independent noise model.

\section{Experiment Protocol}

We run the four learning algorithms on the six robotic tasks to investigate different characteristics such as hyper-parameter and network initialization sensitivities within tasks, hyper-parameter consistency across tasks, and overall learning effectiveness of the algorithms in all tasks.

To analyze the hyper-parameter sensitivity within tasks and consistency across tasks, we perform a random search (Bergstra \& Bengio 2012) of seven hyper-parameters of each algorithm on UR-Reacher-2 and DXL-Reacher.
For each of these hyper-parameters, we predetermine the range of values to search and draw 30 independent hyper-parameter configurations from that range uniformly randomly in the logarithmic scale. 
The ranges of hyper-parameter values we use in this experiment are given in Appendix A.3. 
Each of these hyper-parameter configurations is used to run experiments using a different random initialization of neural networks. 
Each algorithm is run using the same set of hyper-parameter configurations on both tasks.

To know the statistical significance of the comparative performance of each hyper-parameter configuration, we need to run each of them with different randomization seeds that will determine network initialization, random target positions, and random action selections. 
Instead, we run each hyper-parameter configuration of the random search with a single randomly drawn network initialization. 
To determine the effect of the network initialization, we redraw four hyper-parameter configurations uniformly randomly from our original 30 sets. 
Each of these four sets was rerun with five  randomly drawn network initializations. 
We use the same five network initializations for all four sets of chosen hyper-parameter values on both tasks.

To analyze the overall effectiveness of the algorithms across tasks, we choose the best-performing hyper-parameter configurations of each algorithm from UR-Reacher-2 and use them to run experiments on the four held-out tasks: UR-Reacher-6, DXL-Tracker, Create-Mover, and Create-Docker.
To understand the qualitative performance of learned policies, we also run some non-learning scripted agents and compute their average returns using the same experimental setup we use for the learning agents.
These scripted agents are described in Appendix A.4.

Each run is 150,000 steps long or about 3 hours of wall time for UR-Reacher-2, 200,000 steps long or about 4 hours of wall time for UR-Reacher-6, 50,000 steps long or about 45 minutes of wall time for DXL-Reacher, 150,000 steps long or about 2 hours 15 minutes of wall time for DXL-Tracker, 40,000 steps long or about 2 hours of wall time for Create-Mover, and 300,000 steps long for Create-Docker. All wall times include resets. The resets of Create-Docker is dependent on performance.

\section{Experimental results and discussion}

\begin{figure}  
\center
\includegraphics[scale=.155]{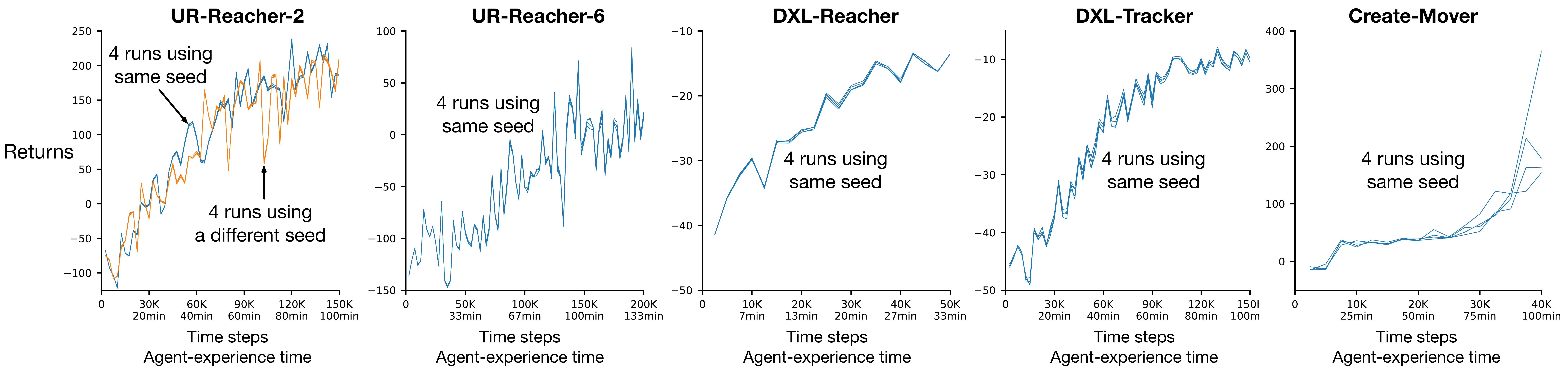}
\caption{
{\bf Repeatability of learning on five robotic tasks:}
The plots show the returns over time of multiple learning experiments that would be identical for the same color if they had been run in simulation.
The robot hardware introduces some non-determinism, but not enough to significantly impact repeatability in the natural ups and downs of exploration and learning, except in Create-Mover, where the physical location of the robot can diverge over time across the runs.
}
\label{fig:repeat}
\end{figure}

\begin{figure}[b] 
\center
\includegraphics[scale=.27]{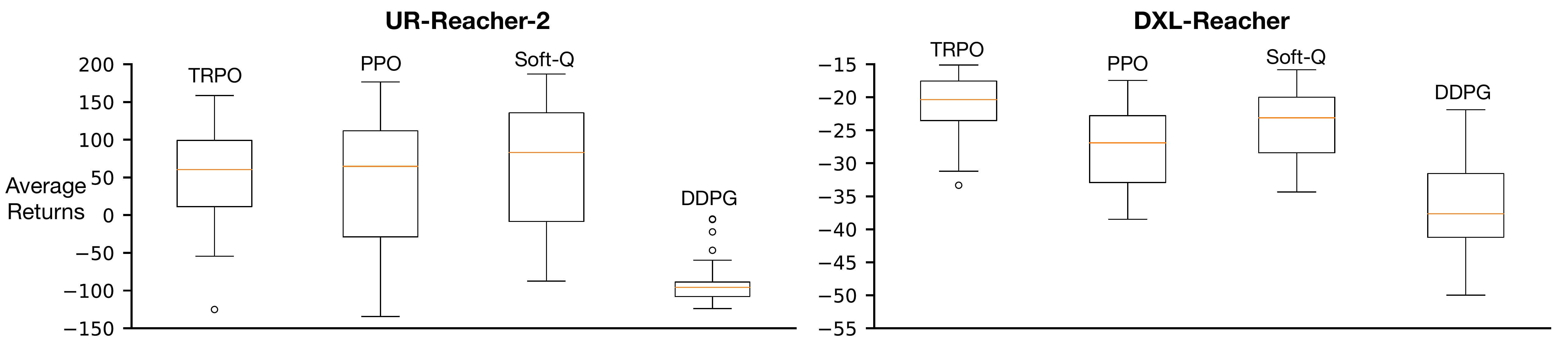}
\caption{
{\bf The effect of hyper-parameter choices in two robotic tasks:}
The plot illustrates the variation in performance due to hyper-parameter choices using boxplots based on 30 randomly drawn hyper-parameter configurations of each algorithm on UR-Reacher-2 and DXL-Reacher tasks.
Hyperparameter choices had a large impact on the quality of learned policies. 
}
\label{fig:boxplot}
\end{figure}

\begin{figure}  
\center
\includegraphics[scale=.19]{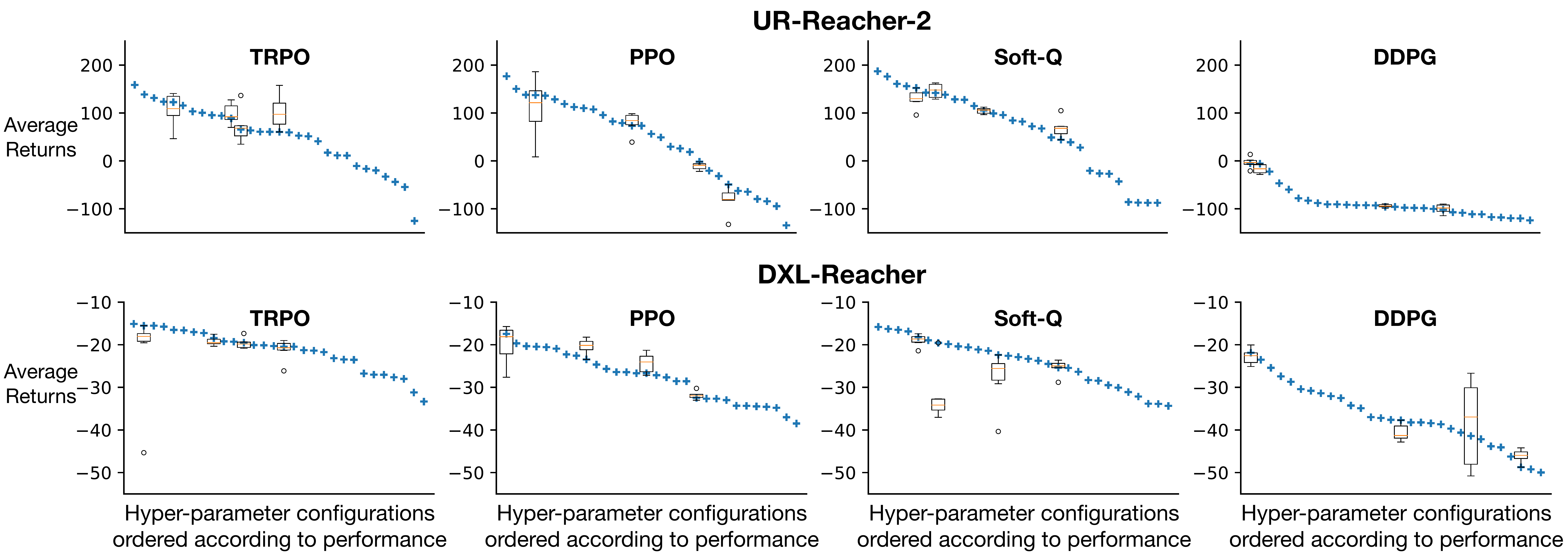}
\caption{
{\bf The effect of random network initializations on DXL-Reacher:}
    Each blue cross represents a single run with a unique hyper-parameter choice. 
They are sorted by performance.
    To quantify the influence of network initialization, we re-ran some of the experiments but varied the initialization.
    The overall effect of network initialization, shown in Tukey's box plots for four randomly chosen hyper-parameter values, was smaller than that of the choice of the hyper-parameters.
}
\label{fig:init}
\end{figure}

\begin{figure}[b]  
\center
\includegraphics[scale=.19]{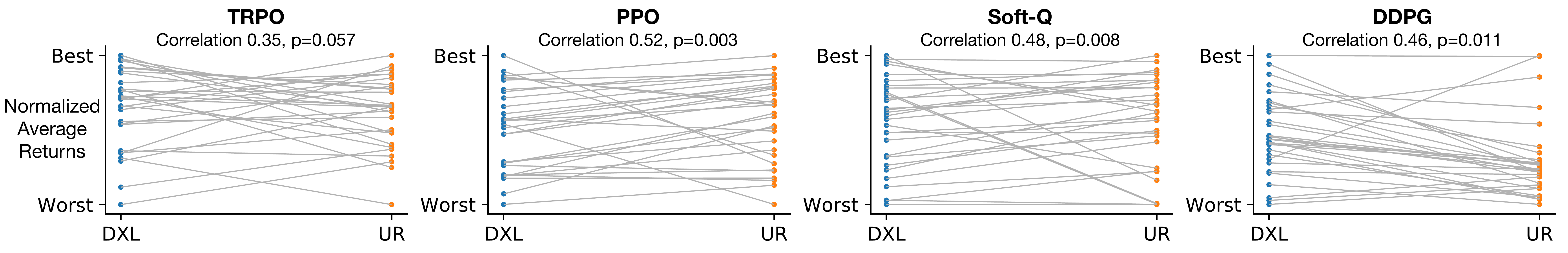}
\caption{
{\bf Hyper-parameter consistency between two robotic tasks:}
Each blue dot represents a single experiment on DXL-Reacher and each orange dot represents an experiment on UR-Reacher-2.
    The gray lines connect experiments with identical hyperparameter choices.
Although the orderings of hyper-parameter configurations by performance were not consistent, all four algorithms had hyper-parameter consistency between the tasks, evident by the correlation of performance shown above each plot, which varied from weakly positive to moderately positive relationships.
}
\label{fig:ordering}
\end{figure}

\begin{figure}  
\center
\includegraphics[scale=.245]{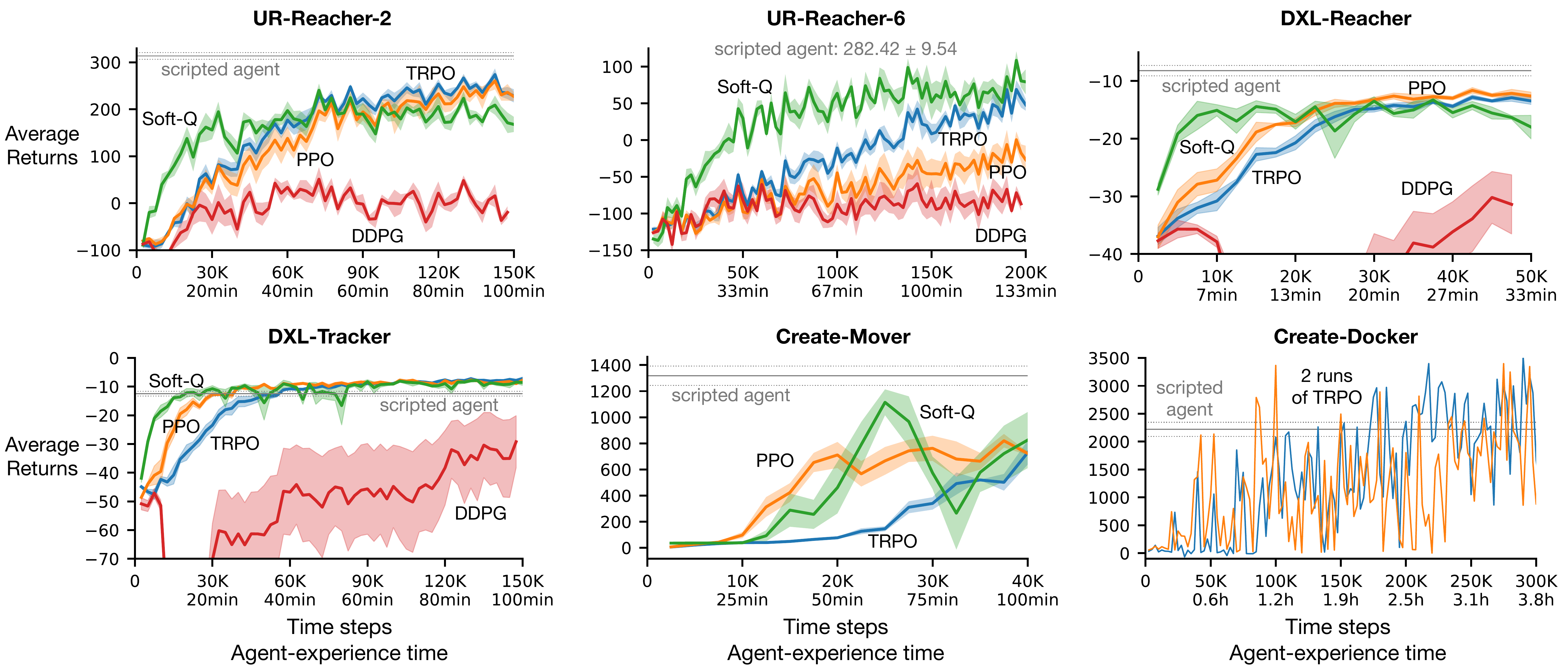}
\caption{
{\bf Learning performance of different algorithms on all six tasks:}
We used the best hyper-parameters based on UR-Reacher-2.
TRPO, PPO and Soft-Q learned effectively, with Soft-Q being the fastest learner and TRPO achieving near-best final learning performance in all tasks.
}
\label{fig:learning}
\end{figure}

First, we demonstrate the reliability of the experiments by repeating some of them multiple times with the same randomization seed using TRPO. Details are given in Appendix A.5.
Figure \ref{fig:repeat} shows the results for all tasks except Create-Docker.
The variation in performance between different runs was small and did not diverge over time except on Create-Mover, where the sequences of experience became dissimilar over time across runs.
These results are a testament to the tight control over system delays achieved in our tasks by using the computational model of Mahmood et al.\ (2018).

To illustrate the sensitivity of each algorithm's performance to their hyper-parameters, we show in Figure \ref{fig:boxplot} the performance of all algorithms on both UR-Reacher-2 (left) and DXL-Reacher (right)  using Tukey's box plots (Tukey 1977) based on all 30 configurations. 
The performance of each configuration was measured by averaging all episodic returns obtained throughout the learning period during its run. 
For each algorithm, performance varied widely with different hyper-parameter configurations, ranging from learning visually-confirmed effective behavior to no learning at all.
No learning occurred for several different hyper-parameter configurations.
The performance of DDPG was the worst, having the least median performance on both tasks.
The rest of the algorithms achieved good performance with many configurations on both tasks. Among them, TRPO's performance was the least sensitive to hyper-parameter variations with the smallest interquartile range on both tasks. 

Overall, these results show that hyper-parameter choices are important, as they may make a much bigger difference than the choice of the algorithm.
Blue crosshairs in Figure \ref{fig:init} show the performance for each of the 30 hyper-parameter configurations of all four algorithms in the descending order, also shown in Appendix A.6 for easier comparison.
The values of all configurations, distributions of individual hyper-parameters and their correlations with performance are given in Appendix A.7.

The box plots in Figure \ref{fig:init} show the effect of variations in network initialization with four randomly chosen hyper-parameter configurations.
Except in one case of DDPG, the interquartile ranges of performance due to variations in network initializations were smaller than those due to variations in hyper-parameter choices shown in Figure \ref{fig:boxplot}.
Except for TRPO on UR-Reacher-2, DDPG on DXL-Reacher and Soft-Q on both tasks, the medians of performance also retained the relative rank order of the configurations in the original experiment with single network initializations.

We show in Figure \ref{fig:ordering} how each hyper-parameter configuration ranks on both tasks according to average returns.
Each plot corresponds to an algorithm and contains two columns of colored dots, each of which represents one of the 30 randomly chosen hyper-parameter configurations. 
The gray lines connect identical hyper-parameter configurations on both tasks.
The correlations of performance between the tasks with corresponding p-values are given above the plots.
The correlations were positive for all algorithms, and significantly large for PPO, Soft-Q and DDPG.
This result indicates that although hyper-parameter optimization is likely necessary for best performance on a new task, a good configuration based on one task can still provide a good baseline performance for another.

Figure \ref{fig:learning} shows the means of multiple learning curves together with their standard errors for different learning algorithms on different tasks, except the bottom right plot, which shows two independent learning curves of TRPO on Create-Docker.
The average returns and standard errors of scripted agents are also shown for each task.
For the mean curves, we used four independent runs of each algorithm on Create-Mover, and five runs on the rest.
DDPG performed poorly on all UR5 and DXL tasks.
We did not run it on the two Create tasks.
The rest of the algorithms showed learning improvements on all tasks they were run.
Among these algorithms, the final performance of TRPO was never substantially worse compared to the best in each task.
Soft-Q had the fastest learning rate on all UR5 and DXL tasks.
On Create-Mover, TRPO, PPO and Soft-Q learned an effective forward-moving behavior of the robot, which turned as it approached the wall, as shown in the companion video\footnote{
A companion video is available at
\url
{https://youtu.be/ovDfhvjpQd8}
}.
On Create-Docker, TRPO learned to dock successfully quite often although the movement was not smooth.
Overall, RL solutions were outperformed by scripted solutions, by a large margin in some tasks, where such solutions were well established or easy to script. 
But in Create-Docker where a scripted solution is not obvious or easy, RL solutions appeared more competitive.

Working with real-world systems did create some challenges.
Soft-Q on DXL tasks for many of its hyper-parameter configurations resulted in frequent overheating and failed during overnight experiments due to more-aggressive exploration.
We could not run un-attended experiments with Create 2 when the robots were tethered to stationary external computers as their cables needed periodic untangling.
We were able to overcome this problem by using an on-board external computer, which we used for one of the two Create-Docker runs.
Two wall sensors of one of the Create 2s were faulty; surprisingly, learning performance did not appear to be affected, possibly due to four other symmetrically oriented wall sensors being available.
Of all three robots, we were most pleased with our experience with UR5, which we were able to use constantly for days without interventions.
Our experiments also revealed some limitations of the learning algorithms and their implementations, such as the sequential computations of the agent's learning updates and forward policy passes.
Learning updates of these algorithms are expensive, and our choice of moderately large action cycle times minimized the number of samples affected by these sequential learning updates.
However, to learn finer policies with faster action cycle times, moving toward efficient ordering of computations (Travnik et al.\ 2018, Mahmood et al.\ 2018) together with inexpensive incremental updates (Mahmood 2017) or asynchronous off-policy updates (Gu et al.\ 2017) would be essential.

\section{Conclusions}

In this work, we provided the first extensive experimental study of multiple policy learning algorithms, namely TRPO, PPO, DDPG, and Soft-Q on multiple commercially-available physical robots.
We found that the performance of all algorithms was highly sensitive to their hyper-parameter values, requiring retuning on new tasks for the best performance.
Nevertheless, some algorithms achieved effective learning performance across tasks for a wide range of hyper-parameter values.
This effectiveness indicates the reliability of our task setups as well as the applicability of these algorithms and implementations in diverse physical environments.
Benchmarking more learning algorithms on these tasks as well as upgrading the existing algorithms to allow higher sample efficiency and faster action cycle times are promising directions for future work.

We ran more than 450 independent experiments which took over 950 hours of robot usage in total.
Most of the experiments were highly repeatable, and many of them resulted in effective learning performance.
This study strongly indicates the viability of reinforcement learning research extensively based on real-world experiments, which is essential to understand the difficulties of learning with physical robots and mitigate them to achieve fast and reliable learning performance in dynamic environments.
The benchmark tasks and the supporting source code enable the necessary steps for such understanding and easy adoption of physical robots in reinforcement learning research.

\section*{Acknowledgements}

We thank Colin Cooke, Francois Hogan, and Daniel Snider for valuable discussion, and Yifei Cheng and Scott Purdy for helping us build the arena for Create 2. Colin Cooke also helped us with the on-board computer setup for one of the two Create-Docker runs.

\newpage

\section*{References}
\newcommand{\hangin}{\goodbreak\hangindent=.15cm \noindent}

\hangin
Benet, G., Blanes, F., Sim\'{o}, J. E., P\'{e}rez, P. (2002). Using infrared sensors for distance measurement in mobile robots. \emph{Robotics and autonomous systems 40}(4), 255--266.

\hangin
Bergstra, J., Bengio, Y. (2012). Random search for hyper-parameter optimization. \emph{Journal of Machine Learning Research}, pp: 281--305.

\hangin
Brockman, G., Cheung, V., Pettersson, L., Schneider, J., Schulman, J., Tang, J., Zaremba, W.\ (2016). OpenAI Gym. \emph{arXiv preprint} arXiv:1606.01540.

\hangin
Dhariwal, P., Hesse, C., Klimov, O., Nichol, A., Plappert, M., Radford, A., Schulman, J., Sidor, S., Wu, Y. (2017). OpenAI Baselines, \url{https://github.com/openai/baselines}.

\hangin
Duan, Y., Chen, X., Houthooft, R., Schulman, J., Abbeel, P.\ (2016). Benchmarking deep reinforcement learning for continuous control. In \emph{Proceedings of the 33rd International Conference on Machine Learning}, pp: 1329--1338.

\hangin
Gu, S., Holly, E., Lillicrap, T., Levine, S.\ (2017). Deep reinforcement learning for robotic manipulation with asynchronous off-policy updates. In \emph{IEEE International Conference on Robotics and Automation}, pp:3389--3396.

\hangin
Haarnoja, T., Tang, H., Abbeel, P., Levine, S. (2017). Reinforcement learning with deep energy-based policies. \emph{arXiv preprint arXiv:1702.08165}.

\hangin
Heess, N., Sriram, S., Lemmon, J., Merel, J., Wayne, G., Tassa, Y., Silver, D. (2017). Emergence of locomotion behaviours in rich environments. \emph{arXiv preprint arXiv:1707.02286}.

\hangin
Henderson, P., Islam, R., Bachman, P., Pineau, J., Precup, D., Meger, D. (2017). Deep reinforcement learning that matters. \emph{arXiv preprint} arXiv:1709.06560.

\hangin
Hester, T., Vecerik, M., Pietquin, O., Lanctot, M., Schaul, T., Piot, B., Horgan, D., Quan, J., Sendonaris, A., Dulac-Arnold, G. and Osband, I., (2017). Deep Q-learning from demonstrations. \emph{arXiv preprint arXiv:1704.03732}.

\hangin
Klissarov, M., Bacon, P. L., Harb, J., Precup, D. (2017). Learnings Options End-to-End for Continuous Action Tasks. \emph{arXiv preprint arXiv:1712.00004}.

\hangin
Levine, S., Finn, C., Darrell, T., Abbeel, P.~(2016). End-to-end training of deep visuomotor policies. \emph{The Journal of Machine Learning Research 17}(1): 1334--1373.

\hangin
Lillicrap, T.P., Hunt, J.J., Pritzel, A., Heess, N., Erez, T., Tassa, Y., Silver, D. and Wierstra, D. (2015). Continuous control with deep reinforcement learning. \emph{arXiv preprint arXiv:1509.02971}.

\hangin

Machado, M. C., Bellemare, M. G.,  Talvitie, E., Veness, J., Hausknecht, M., Bowling, M.~(2018). Revisiting the arcade learning environment: Evaluation protocols and ppen problems for general agents. \emph{Journal of Artificial Intelligence Research 61}: 523--562.

\hangin
Mahmood, A. R. (2017). \emph{Incremental Off-policy Reinforcement Learning Algorithms}. PhD thesis, Department of Computing Science, University of Alberta, Edmonton, AB T6G 2E8.

\hangin
Mahmood, A. R., Korenkevych, D., Komer, B. J., Bergstra, J. (2018). Setting up a Reinforcement Learning Task with a Real-World Robot. \emph{arXiv preprint arXiv:1803.07067}.

\hangin
Riedmiller, M. (2012). 10 steps and some tricks to set up neural reinforcement controllers. In \emph{Neural networks: tricks of the trade}, pp. 735--757. Springer, Berlin, Heidelberg.

\hangin
Riedmiller, M., Hafner, R., Lampe, T., Neunert, M., Degrave, J., Van de Wiele, T., Mnih, V., Heess, N. and Springenberg, J.~T., (2018). Learning by Playing-Solving Sparse Reward Tasks from Scratch. arXiv preprint arXiv:1802.10567.

\hangin
Rusu, A.\ A., Ve\u cer\' ik, M., Roth\" orl, T., Heess, N., Pascanu, R., Hadsell, R.\ (2017). Sim-to-Real robot learning from pixels with progressive nets. In \emph{Proceedings of the 2nd Conference on Robot Learning}, pp: 262--270.

\hangin
Schulman, J., Levine, S., Abbeel, P., Jordan, M., Moritz, P.\ (2015). Trust region policy optimization. In \emph{Proceedings of the 32nd International Conference on Machine Learning}, pp:1889--1897.

\hangin
Schulman, J., Wolski, F., Dhariwal, P., Radford, A., Klimov, O. (2017). Proximal policy optimization algorithms. \emph{arXiv preprint arXiv:1707.06347}.

\hangin
Sutton, R.\ S., Barto, A.\ G.\ (1998). \emph{Reinforcement Learning: An Introduction}. MIT Press.

\hangin
Tassa, Y., Doron, Y., Muldal, A., Erez, T., Li, Y., de Las Casas, D., Budden, D., Abdolmaleki, A., Merel, J., Lefrancq, A., Lillicrap, T., and Riedmiller, M.~(2018). DeepMind control suite. \emph{arXiv preprint arXiv:1801.00690}.

\hangin
Travnik, J. B., Mathewson, K. W., Sutton, R. S., Pilarski, P. M. (2018). Reactive Reinforcement Learning in Asynchronous Environments. \emph{arXiv preprint arXiv:1802.06139}.

\hangin
Tukey, J. W. (1977). Box-and-whisker plots. \emph{Exploratory data analysis}, pp: 39--43.

\hangin
Wulfmeier, M., Posner, I., Abbeel, P. (2017). Mutual alignment transfer learning. \emph{arXiv preprint arXiv:1707.07907}.

\hangin
Yahya, A., Li, A., Kalakrishnan, M., Chebotar, Y., Levine, S. (2017). Collective robot reinforcement learning with distributed asynchronous guided policy search. In \emph{2017 IEEE/RSJ International Conference on Intelligent Robots and Systems}, pp: 79--86.

\newpage

\section*{Appendix} 

\section*{A.1 Additional details of the robots}

Here, we provide additional details on the hardware setup between the robots and the control computers. 
All of our setups use wired connections.
The UR5 arm controller is communicated with the control computer over a TCP/IP connection.
We use an Xevelabs USB2AX controller to interface between the MX-64AT actuators and a control computer via USB.
A 12V, 5A DC power adapter is used to power the actuators. 
The Create 2 robot is interfaced with a control computer via serial port using iRobot's specified Open Interface. The robot is communicated in the streaming mode where the internal controller streams a data packet every 15ms, which is the rate the internal controller uses to update data.

\section*{A.2 Additional details of Create-Docker}

In Create-Docker, the objective is to dock to a charging station attached to the middle of one of the wider walls of the Create-Mover arena. 
When the robot is at the charging station in such a way that the binary charging signal is active, the internal robot controller switches the operating mode to \emph{Passive}, in which the actuation commands for the wheels from the external computer are ignored.
Being in this state with an active charging signal is considered a successful docking.
The internal controller does not switch to the Passive mode right away after an active charging signal.
Therefore, it is possible to have an active charging signal momentarily but still not dock successfully due to a high speed in the backward direction or bouncing back from the station.
Moreover, it is extremely difficult to activate the charging signal properly if the robot approaches the charging station at an angle.
Therefore, to learn how to dock, it is important to approach the charging station perpendicularly and learn to slow down or stop when the charging signal is active.

The action space for Create-Docker is the same as for Create-Mover, that is, $[-150mm/s, 150mm/s]^2$ for actuating the two wheels with speed control.
The observation vector is 20-dimensional, consisting of a single binary charging signal, six infrared wall signals, two bump signals, two action components from the previous time step, and nine additional components processed based on the three infrared bytes from the charging station. 
Each infrared byte informs whether the left buoy, the right buoy and the force field of the dock beam configuration can be seen, from which we obtain nine binary values. 
Each of these binary values are then averaged over last 20 packets streamed from the robot's internal controller every 15ms.

The reward function is a large positive number for successful docking with penalty for bumping and encouragement for moving forward and facing the charging station perpendicularly. 
The reward function for every time-step is defined as follows:
\begin{align}
R_t &= \tau \left( a X_t + b Y_t + c Z_t + d V_t \right),
\end{align}
where $\tau$ is the action cycle time, $X_t$ is the docking reward, $Y_t$ is bumping penalty, $Z_t$ is the moving bonus and $V_t$ is the bonus for aligning to the charging station. These components are defined as follows:
\begin{align}
X_t &= \frac{2}{n(n+1)} \sum_{i=1}^{n} (n-i+1)p_t(charging, i) \\
Y_t &= - \sum_{k=1}^2 \bigcup_{i=1}^n p_t(bump_{k}, i) \\
Z_t &= \frac{1}{n} \sum_{i=1}^n p_t(distance, i) \\
V_t &= \frac{1}{20} \sum_{i=1}^{20} \sum_{k=1}^{9} w_k p_t[ir\_dock_k, i].
\end{align}
Here, $p_t[x, i]$ stands for the $i$th most-recent data packet available at time step $t$ for sensor $x$, where $charging$ stands for charging sensor, $bump_k$ stands for $k$th bump sensor, $distance$ stands for the distance sensor and $ir\_dock_k$ stands for the $k$th sensor value for the infrared receiver for docking. The weights used for $ir\_dock$ are $w = [1.0, 0.5, 0.05, 0.65, 0.15, 0.65, 0.05, 0.5, 1.0]$, which are chosen in such a way that the left receiver (first three values) focuses on the left buoy, the right receiver (last three values) focuses on the right buoy and the omni-directional receiver (middle three values) focuses on both buoy equally. The value of $n$ is the ratio between action cycle time and the read-write cycle time, that is, $n = \frac{\tau}{0.015}$. The weights of the different reward components are $a=150$, $b=10$, $c=5$, and $d=4$. They are chosen in such a way that the maximums of the penalty and the two bonuses are of the same magnitude scale and the docking reward is much larger than the auxiliary reward components.
If docking is successful, the robot stays at the charging station for the rest of the episode and continues to receive the docking reward.
This encourages docking as soon as possible.

An episode is always 30 seconds long. 
We designed the reset between episodes in such a way that docking is relatively easier for the initial policy if the previous episode is unsuccessful and it is relatively more difficult if the previous episode is successful, that is, the episode terminated while the robot is docked. 
To achieve this, the reset procedure first invokes the internal \emph{seek-dock} routine of Open Interface to dock to the station if the previous episode is unsuccessful. 
After the robot docks using seek-dock or a time-out of 20 seconds, the robot moves backward for 3.25 seconds. 
If the seek-dock succeeds, the robot is always facing the charging station after reset and can dock successfully by learning to move straight in the forward direction. 
However, if the seek-dock routine does not succeed, then the robot may start from a difficult starting position, for example, at one of the corners of the arena facing the wall.
If the previous episode is successful, then the reset procedure makes the robot move backward for 0.75 seconds and then sends uniform random speeds for 2.5 seconds to the two wheels independently between $[-250, -50]$ to move backward further rotationally. 
This last phase ensures that the robot is likely not facing the charging station perpendicularly and displacement is required to achieve alignment.

\section*{A.3 Ranges of hyper-parameter values for random search}

\subsection*{TRPO:}

\begin{center}
\begin{tabular}{ |c|c| } 
 \hline
 Hyperparameter & Range \\ 
 \hline 
 batch size & $2^{[8, 13]}$ \\
 \rowcolor{Gray}
 vf-step-size & $10^{[-5, -2]}$ \\ 
 $\delta_{KL}$ & $10^{[-2.5, -0.5]}$ \\
  \rowcolor{Gray}
 $c_{\gamma}$ & $10^{[\log_{10}(10/N), 1.5]}$ \\
  $\gamma $ & $1-\frac{1}{c_{\gamma}N}$ \\
\rowcolor{Gray}
 $c_{\lambda}$ & $10^{[\log_{10}(10/N), 1.5]}$ \\
$\lambda $ & $1-\frac{1}{c_{\lambda}N}$ \\
 \rowcolor{Gray}
 hidden layers & $[1, 4]$ \\
 hidden sizes & $2^{[3, X ]}$ \\
 \hline
\end{tabular}
\end{center}
Here, $N = T/\tau$, where $T$ is the total length of an episode in time and $\tau$ is the action cycle time.
We restricted total number of weights in the network to be no larger than 100,000, and the upper limit of a hidden size $X$ was determined based on sampled number of hidden layers to respect this limit.

\subsection*{PPO:}

\begin{center}
\begin{tabular}{ |c|c| } 
 \hline 
 Hyperparameter & Range \\ 
 \hline 
 batch size & $2^{[8, 13]}$ \\
 \rowcolor{Gray}
 step-size & $10^{[-5, -2]}$ \\ 
 opt. batch size & $2^{[3,\ \log_2 (\text{batch size})]}$ \\
  \rowcolor{Gray}
 $c_{\gamma}$ & $10^{[\log_{10}(10/N), 1.5]}$ \\
  $\gamma $ & $1-\frac{1}{c_{\gamma}N}$ \\
\rowcolor{Gray}
 $c_{\lambda}$ & $10^{[\log_{10}(10/N), 1.5]}$ \\
$\lambda $ & $1-\frac{1}{c_{\lambda}N}$ \\
 \rowcolor{Gray}
 hidden layers & $[1, 4]$ \\
 hidden sizes & $2^{[3, X]}$ \\
 \hline
\end{tabular}
\end{center}

\subsection*{Soft-Q:}

\begin{center}
\begin{tabular}{ |c|c| } 
 \hline 
 Hyperparameter & Range \\ 
 \hline 
 batch size & $2^{[8, 13]}$ \\
 \rowcolor{Gray}
 step size & $10^{[-5, -2]}$ \\
  epochs & $2^{[0,\ 2]}$ \\  
  \rowcolor{Gray}
 $c_{\gamma}$ & $10^{[\log_{10}(10/N), 1.5]}$ \\
  $\gamma $ & $1-\frac{1}{c_{\gamma}N}$ \\
\rowcolor{Gray}
  reward scale & $10^{[0,\ 2]}$ \\
  hidden layers & $[1, 4]$ \\
\rowcolor{Gray}
 hidden sizes & $2^{[3, X]}$ \\
 \hline
\end{tabular}
\end{center}

\subsection*{DDPG:}

\begin{center}
\begin{tabular}{ |c|c| } 
 \hline 
 Hyperparameter & Range \\ 
 \hline 
 batch size & $2^{[8, 13]}$ \\
 \rowcolor{Gray}
 step size & $10^{[-5, -2]}$ \\
  exploration $\sigma$ & $10^{[-2,\ \log_{10}5]}$ \\  
  \rowcolor{Gray}
 $c_{\gamma}$ & $10^{[\log_{10}(10/N), 1.5]}$ \\
  $\gamma $ & $1-\frac{1}{c_{\gamma}N}$ \\
 \rowcolor{Gray}
 reward scale & $10^{[0,\ 2]}$ \\
 hidden layers & $[1, 4]$ \\
 \rowcolor{Gray}
 hidden sizes & $2^{[3, X]}$ \\
 \hline
\end{tabular}
\end{center}

~

~

\section*{A.4 Details of scripted agents}

For UR5 tasks, we use the \emph{movej} command of URScript, where we specify the joint angles for the target positions and set the time to reach to 2 seconds. 
For DXL tasks, we implement a PID controller. We do not constrain the current control values as we do for the learning agent, and we chose the optimal PID gain value for each task separately.
For Create-Mover, we use a simple script that applies action $[-150mm/s, +150mm/s]$ whenever the normalized signal value of either of the two front wall sensors has its value above 0.55 and  otherwise, moves straight forward with action $[+150mm/s, +150mm/s]$. 
For Create-Docker, we use the \emph{seek-dock} routine of Open Interface. During this routine, the robot moves back and forth to perceive the vertical plane perpendicular to the wall at the charging station using the infrared bytes from the charging station, adjusts its position to align its principle axis with the perpendicular plane, and moves slowly toward the charging station.

\section*{A.5 Details of repeatability experiments with TRPO}

To generate the plots in Figure 2, we run the same experiment with the same seed four times on five different tasks using TRPO. There are different kinds of randomization in each task. For the agent, randomization is used to initialize the network and sample actions. For the environment, randomization is used to generate targets and resets. By using the same randomization seed across multiple experiments in this set of experiments, we ensure that the environment generates the same sequence of targets and resets, the agent is initialized with the same network, and it generates the same or similar sequence of actions for a particular task. We use the same hyper-parameter values of TRPO used in the experiments by Henderson et al.\ (2017).

\newpage

\section*{A.6 Relative performance of different algorithms in random search}

In the figure below, we show the relative performance of four algorithms across 30 random hyper-parameters configurations ordered by performance. 
Note that for different algorithms, parameter configurations with the same index generally correspond to different parameter values.  

\begin{figure}[H]
\includegraphics[scale=0.4]{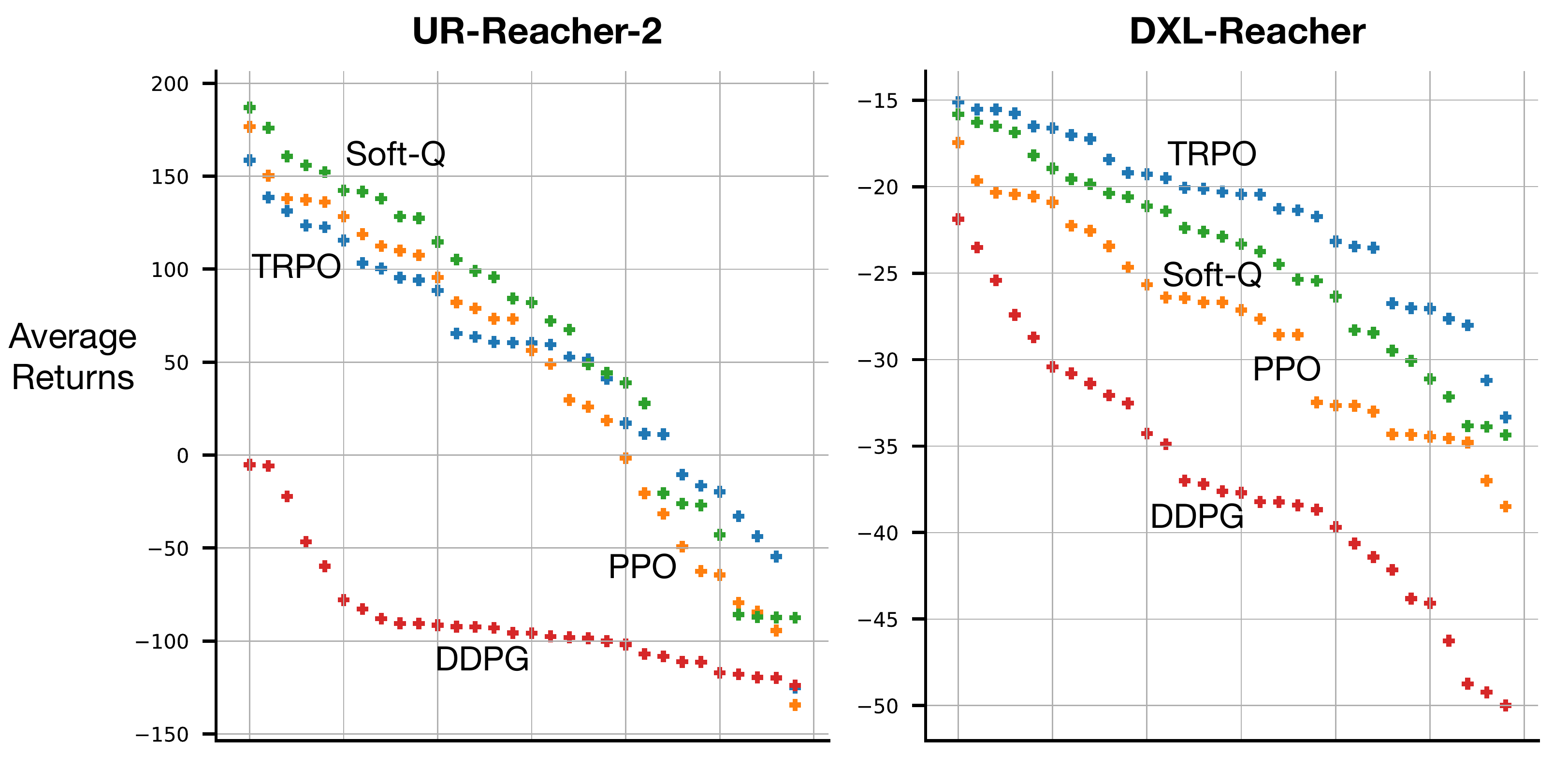}
\caption{
All random parameter configurations of all four algorithms on UR-Reacher-2 and DXL-Reacher.
}
\label{fig:ur5_ppo_params}
\end{figure}

\newpage

\section*{A.7 All hyper-parameter configurations, their value distributions and correlations with returns}

In the following table, we show parameter values for all 30 configurations and their corresponding average returns for TRPO on UR-Reacher-2. The configurations are shown in descending order according to average returns.

\begin{table}[h]
\begin{center}
\begin{tabular}{ |c|c|c|c|c|c|c|c|c|c| } 
 \hline
 Average & batch & vf-step-size & $\delta_{KL}$ & $\gamma$ & $\lambda$ & hidden & hidden \\
 Return & size & & & & & layers & sizes \\
\hline
158.56 &  4096 &  0.00472 &  0.02437 &  0.96833 &  0.99874 &     2 &    64 \\ 
\rowcolor{Gray}
  138.58 &  2048 &  0.00475 &  0.01909 &  0.99924 &  0.99003 &     1 &   128 \\ 
  131.35 &  8192 &  0.00037 &  0.31222 &  0.97433 &  0.99647 &     4 &    64 \\ 
\rowcolor{Gray}
  123.45 &  4096 &  0.00036 &  0.01952 &  0.99799 &  0.92958 &     4 &   128 \\ 
  122.60 &  2048 &  0.00163 &  0.00510 &  0.96801 &  0.96893 &     4 &    32 \\ 
\rowcolor{Gray}
  115.51 &  4096 &  0.00926 &  0.01659 &  0.99935 &  0.99711 &     3 &     8 \\ 
  103.18 &  4096 &  0.00005 &  0.21515 &  0.99891 &  0.99880 &     1 &     8 \\ 
\rowcolor{Gray}
  100.38 &  8192 &  0.00005 &  0.09138 &  0.99677 &  0.99959 &     1 &    64 \\ 
   95.47 &  2048 &  0.00001 &  0.06088 &  0.98488 &  0.99957 &     2 &   128 \\ 
\rowcolor{Gray}
   94.16 &  2048 &  0.00770 &  0.02278 &  0.99414 &  0.98684 &     3 &   128 \\ 
   88.57 &  4096 &  0.00282 &  0.02312 &  0.99813 &  0.99964 &     4 &    16 \\ 
\rowcolor{Gray}
   65.44 &   512 &  0.00054 &  0.01882 &  0.99728 &  0.99420 &     3 &    32 \\ 
   63.60 &  8192 &  0.00009 &  0.10678 &  0.97415 &  0.99759 &     2 &   128 \\ 
\rowcolor{Gray}
   60.79 &  1024 &  0.00007 &  0.02759 &  0.99945 &  0.99961 &     3 &     8 \\ 
   60.51 &  4096 &  0.00222 &  0.00392 &  0.98544 &  0.98067 &     4 &     8 \\ 
\rowcolor{Gray}
   60.35 &  8192 &  0.00004 &  0.25681 &  0.99750 &  0.98955 &     2 &   128 \\ 
   59.39 &  1024 &  0.00435 &  0.00518 &  0.99516 &  0.99867 &     4 &    32 \\ 
\rowcolor{Gray}
   52.70 &  8192 &  0.00001 &  0.03385 &  0.99119 &  0.98400 &     4 &    32 \\ 
   51.44 &   512 &  0.00034 &  0.01319 &  0.97334 &  0.98524 &     4 &    16 \\ 
\rowcolor{Gray}
   41.05 &   512 &  0.00001 &  0.00351 &  0.99430 &  0.99781 &     3 &     8 \\ 
   17.14 &  8192 &  0.00023 &  0.01305 &  0.95963 &  0.99950 &     3 &    32 \\ 
\rowcolor{Gray}
   11.43 &   512 &  0.00251 &  0.00532 &  0.99447 &  0.99951 &     3 &    64 \\ 
   11.13 &   512 &  0.00003 &  0.00727 &  0.99686 &  0.93165 &     1 &   256 \\ 
\rowcolor{Gray}
  -10.57 &   256 &  0.00065 &  0.04867 &  0.99926 &  0.98226 &     1 &    16 \\ 
  -16.48 &  8192 &  0.00001 &  0.31390 &  0.99948 &  0.99204 &     2 &    16 \\ 
\rowcolor{Gray}
  -19.78 &   512 &  0.00005 &  0.15077 &  0.96836 &  0.99944 &     3 &    64 \\ 
  -32.85 &   256 &  0.00003 &  0.12650 &  0.99260 &  0.98021 &     4 &   128 \\ 
\rowcolor{Gray}
  -43.74 &  8192 &  0.00018 &  0.00333 &  0.98940 &  0.97090 &     3 &     8 \\ 
  -54.55 &   512 &  0.00011 &  0.07420 &  0.99402 &  0.90185 &     1 &  2048 \\ 
\rowcolor{Gray}
 -125.13 &   256 &  0.00002 &  0.05471 &  0.99961 &  0.99877 &     4 &    32 \\
 \hline 
\end{tabular}
\end{center}
\caption{
The parameter values and corresponding average returns for all 30 hyper-parameter configurations of TRPO on UR-Reacher-2.
}
\end{table}

\newpage
In the figure below, we show the best 5 (in red) and the worst 5 (in blue) hyper-parameter values out of 30 random configurations  of TRPO on UR-Reacher-2. 
On each plot the x axis represents parameter values and the y axis represents average returns obtained during the corresponding run. 
For each hyper-parameter, we also show correlations between log-parameter values and corresponding returns. 

\begin{figure}[h]
\includegraphics[scale=0.45]{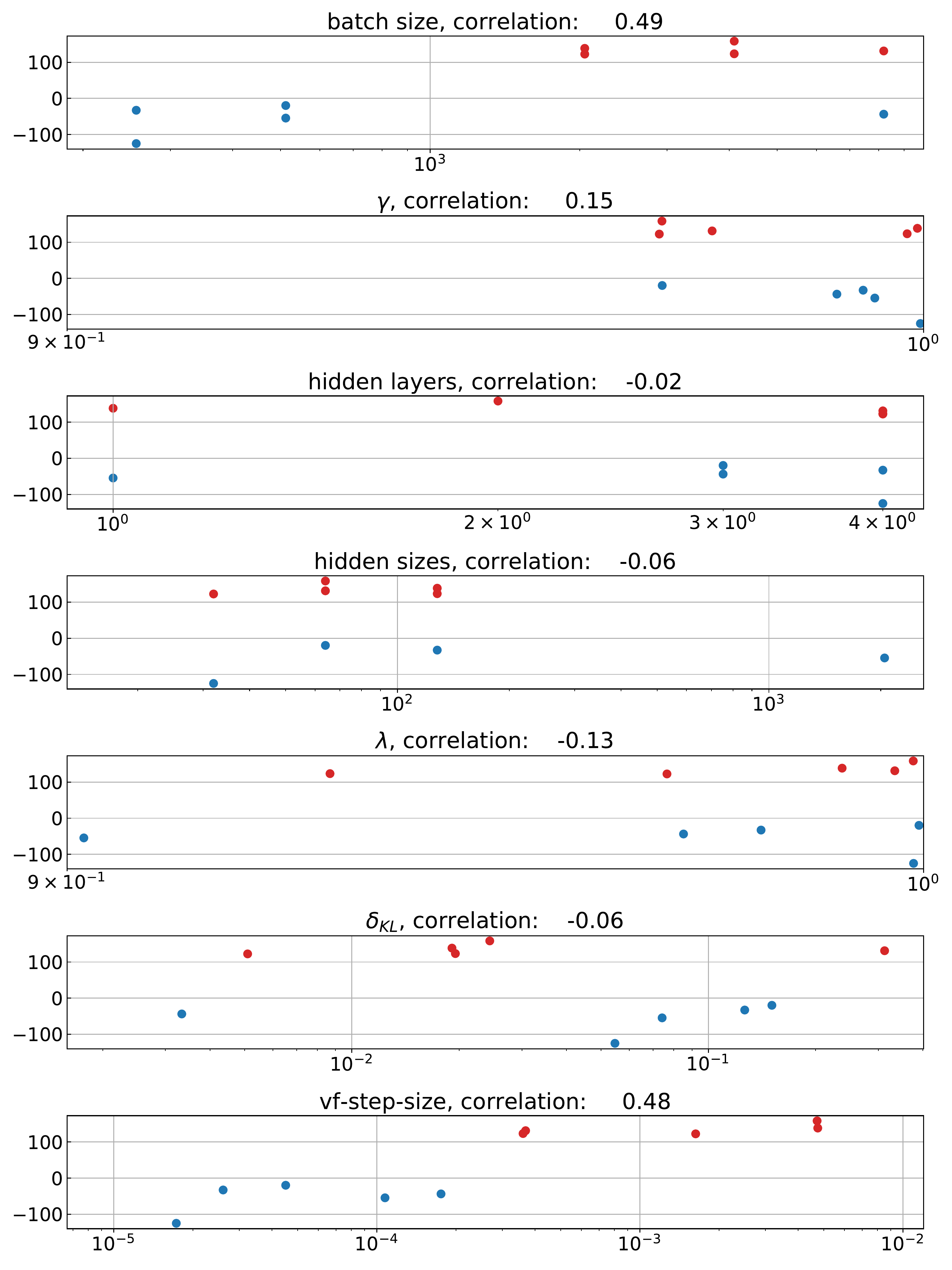}
\caption{
Average returns (y axis) vs. parameter values (x axis) for
best 5 and worst 5 hyper-parameter configurations of TRPO on UR-Reacher-2.}
\label{fig:ur5_ppo_params}
\end{figure}

\newpage

\begin{table}[h]
\begin{center}
\begin{tabular}{ |c|c|c|c|c|c|c|c| } 
 \hline
 Average & batch & vf-step-size & $\delta_{KL}$ & $\gamma$ & $\lambda$ & hidden & hidden \\
 Return & size & & & & & layers & sizes \\
 \hline
 -15.11 &   512 &  0.00251 &  0.00532 &  0.99216 &  0.99907 &     3 &    64 \\ 
\rowcolor{Gray}
  -15.53 &   512 &  0.00054 &  0.01882 &  0.99580 &  0.99183 &     3 &    32 \\ 
  -15.54 &   256 &  0.00065 &  0.04867 &  0.99867 &  0.97815 &     1 &    16 \\ 
\rowcolor{Gray}
  -15.76 &   512 &  0.00034 &  0.01319 &  0.96874 &  0.98142 &     4 &    16 \\ 
  -16.52 &  2048 &  0.00001 &  0.06088 &  0.98102 &  0.99918 &     2 &   128 \\ 
\rowcolor{Gray}
  -16.61 &  2048 &  0.00770 &  0.02278 &  0.99175 &  0.98320 &     3 &   128 \\ 
  -17.03 &  4096 &  0.00005 &  0.21515 &  0.99812 &  0.99796 &     1 &     8 \\ 
\rowcolor{Gray}
  -17.24 &  1024 &  0.00435 &  0.00518 &  0.99304 &  0.99777 &     4 &    32 \\ 
  -18.44 &  2048 &  0.00163 &  0.00510 &  0.96331 &  0.96423 &     4 &    32 \\ 
\rowcolor{Gray}
  -19.21 &  1024 &  0.00007 &  0.02759 &  0.99897 &  0.99924 &     3 &     8 \\ 
  -19.29 &   512 &  0.00011 &  0.07420 &  0.99161 &  0.90163 &     1 &  2048 \\ 
\rowcolor{Gray}
  -19.51 &  4096 &  0.00282 &  0.02312 &  0.99699 &  0.99929 &     4 &    16 \\ 
  -20.07 &  8192 &  0.00009 &  0.10678 &  0.96958 &  0.99622 &     2 &   128 \\ 
\rowcolor{Gray}
  -20.12 &  8192 &  0.00001 &  0.31390 &  0.99902 &  0.98921 &     2 &    16 \\ 
  -20.30 &  4096 &  0.00036 &  0.01952 &  0.99678 &  0.92655 &     4 &   128 \\ 
\rowcolor{Gray}
  -20.45 &  8192 &  0.00004 &  0.25681 &  0.99610 &  0.98628 &     2 &   128 \\ 
  -20.45 &  4096 &  0.00472 &  0.02437 &  0.96363 &  0.99786 &     2 &    64 \\ 
\rowcolor{Gray}
  -21.28 &   512 &  0.00003 &  0.00727 &  0.99524 &  0.92844 &     1 &   256 \\ 
  -21.37 &  8192 &  0.00037 &  0.31222 &  0.96976 &  0.99472 &     4 &    64 \\ 
\rowcolor{Gray}
  -21.73 &  4096 &  0.00926 &  0.01659 &  0.99880 &  0.99558 &     3 &     8 \\ 
  -23.17 &   512 &  0.00001 &  0.00351 &  0.99195 &  0.99653 &     3 &     8 \\ 
\rowcolor{Gray}
  -23.47 &  8192 &  0.00005 &  0.09138 &  0.99512 &  0.99921 &     1 &    64 \\ 
  -23.54 &  8192 &  0.00001 &  0.03385 &  0.98820 &  0.98005 &     4 &    32 \\ 
\rowcolor{Gray}
  -26.75 &   256 &  0.00003 &  0.12650 &  0.98987 &  0.97595 &     4 &   128 \\ 
  -27.02 &  8192 &  0.00023 &  0.01305 &  0.95497 &  0.99905 &     3 &    32 \\ 
\rowcolor{Gray}
  -27.05 &  2048 &  0.00475 &  0.01909 &  0.99864 &  0.98684 &     1 &   128 \\ 
  -27.64 &   256 &  0.00002 &  0.05471 &  0.99924 &  0.99792 &     4 &    32 \\ 
\rowcolor{Gray}
  -28.02 &  4096 &  0.00222 &  0.00392 &  0.98164 &  0.97644 &     4 &     8 \\ 
  -31.20 &   512 &  0.00005 &  0.15077 &  0.96366 &  0.99895 &     3 &    64 \\ 
\rowcolor{Gray}
  -33.33 &  8192 &  0.00018 &  0.00333 &  0.98611 &  0.96624 &     3 &     8 \\ 
 \hline 
\end{tabular}
\end{center}
\caption{
The parameter values and corresponding average returns for all 30 hyper-parameter configurations of TRPO on DXL-Reacher.
}
\end{table}

\newpage

\begin{figure}[h]
\includegraphics[scale=0.45]{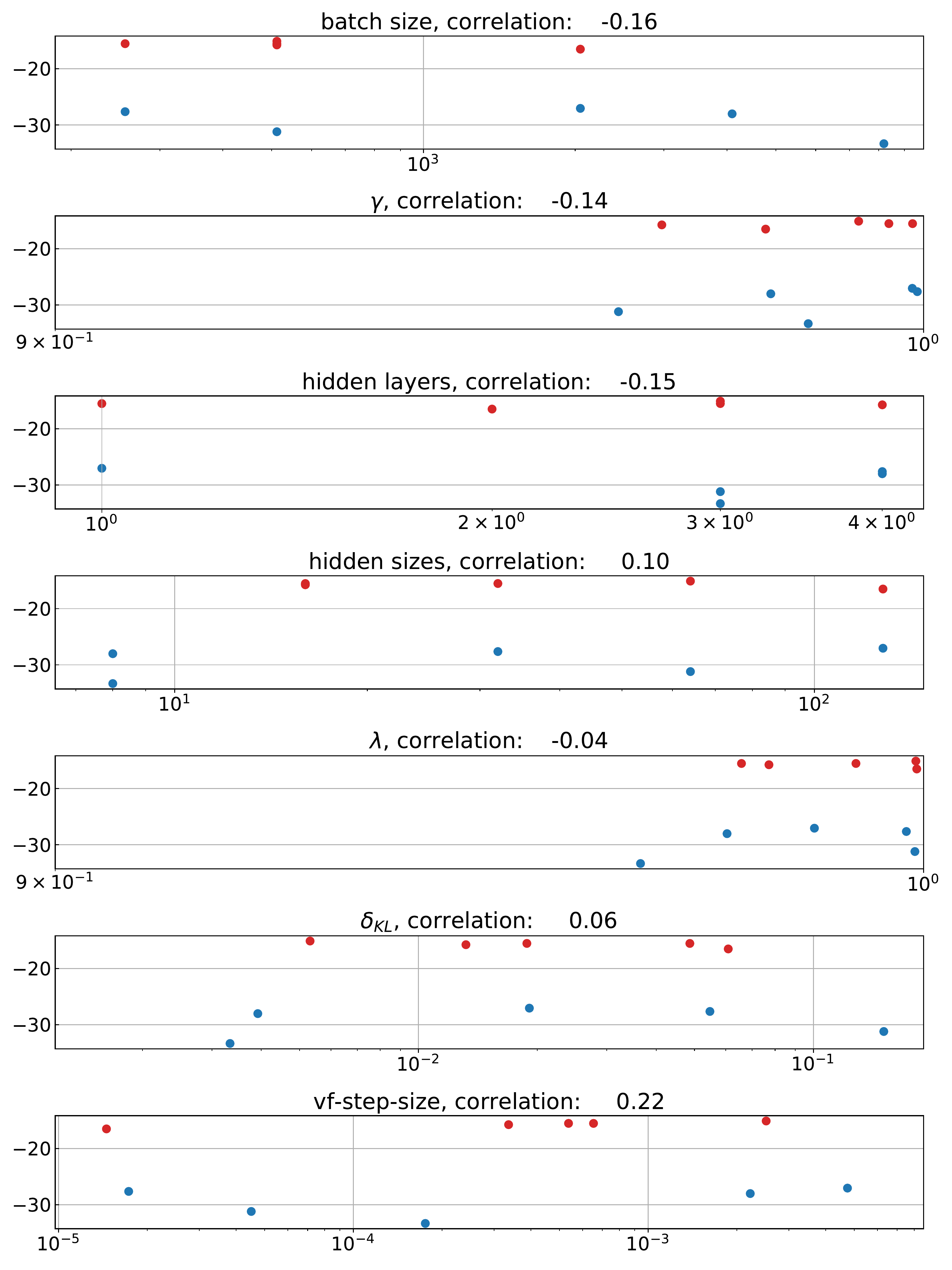}
\caption{
Average return (y axis) vs. parameter values (x axis) for
best 5 and worst 5 hyper-parameter configurations of TRPO on  DXL-Reacher.}
\label{fig:ur5_ppo_params}
\end{figure}

\newpage

\begin{table}[h]
\begin{center}
\begin{tabular}{ |c|c|c|c|c|c|c|c| } 
 \hline
 Average & batch & step-size & opt. & $\gamma$ & $\lambda$ & hidden & hidden \\
 Return &size & & batch size & & & layers & sizes \\
\hline
 176.62 &   512 &  0.00005 &    16 &  0.96836 &  0.99944 &     3 &    64 \\ 
\rowcolor{Gray}
  150.25 &   256 &  0.00050 &    64 &  0.99926 &  0.98226 &     1 &    16 \\ 
  137.92 &   512 &  0.00011 &     8 &  0.99402 &  0.90185 &     1 &  2048 \\ 
\rowcolor{Gray}
  137.26 &  2048 &  0.00163 &  1024 &  0.96801 &  0.96893 &     4 &    32 \\ 
  136.09 &  2048 &  0.00280 &    32 &  0.99924 &  0.99003 &     1 &   128 \\ 
\rowcolor{Gray}
  128.34 &  4096 &  0.00036 &    64 &  0.99799 &  0.92958 &     4 &   128 \\ 
  118.77 &   512 &  0.00003 &    32 &  0.99686 &  0.93165 &     1 &   256 \\ 
\rowcolor{Gray}
  112.48 &  4096 &  0.00941 &  1024 &  0.98544 &  0.98067 &     4 &     8 \\ 
  110.01 &  4096 &  0.00080 &     8 &  0.99935 &  0.99711 &     3 &     8 \\ 
\rowcolor{Gray}
  107.47 &  4096 &  0.00267 &  4096 &  0.96833 &  0.99874 &     2 &    64 \\ 
   95.62 &  8192 &  0.00226 &    32 &  0.97433 &  0.99647 &     4 &    64 \\ 
\rowcolor{Gray}
   82.21 &  8192 &  0.00037 &    16 &  0.99119 &  0.98400 &     4 &    32 \\ 
   78.97 &   512 &  0.00090 &   128 &  0.99430 &  0.99781 &     3 &     8 \\ 
\rowcolor{Gray}
   73.33 &  4096 &  0.00079 &    32 &  0.99813 &  0.99964 &     4 &    16 \\ 
   73.17 &   256 &  0.00003 &    16 &  0.99260 &  0.98021 &     4 &   128 \\ 
\rowcolor{Gray}
   56.25 &  8192 &  0.00987 &    32 &  0.99948 &  0.99204 &     2 &    16 \\ 
   49.02 &  8192 &  0.00019 &    64 &  0.99677 &  0.99959 &     1 &    64 \\ 
\rowcolor{Gray}
   29.70 &  1024 &  0.00039 &   256 &  0.99945 &  0.99961 &     3 &     8 \\ 
   25.94 &  8192 &  0.00362 &    32 &  0.97415 &  0.99759 &     2 &   128 \\ 
\rowcolor{Gray}
   18.64 &  4096 &  0.00061 &   512 &  0.99891 &  0.99880 &     1 &     8 \\ 
   -1.68 &  8192 &  0.00006 &  2048 &  0.99750 &  0.98955 &     2 &   128 \\ 
\rowcolor{Gray}
  -20.53 &  8192 &  0.00087 &  2048 &  0.98940 &  0.97090 &     3 &     8 \\ 
  -31.58 &  1024 &  0.00315 &   256 &  0.99516 &  0.99867 &     4 &    32 \\ 
\rowcolor{Gray}
  -49.32 &   512 &  0.00680 &   128 &  0.99728 &  0.99420 &     3 &    32 \\ 
  -62.46 &  2048 &  0.00002 &  2048 &  0.99414 &  0.98684 &     3 &   128 \\ 
\rowcolor{Gray}
  -64.50 &  2048 &  0.00002 &  1024 &  0.98488 &  0.99957 &     2 &   128 \\ 
  -79.45 &  8192 &  0.00002 &   256 &  0.95963 &  0.99950 &     3 &    32 \\ 
\rowcolor{Gray}
  -84.29 &   512 &  0.00002 &   256 &  0.97334 &  0.98524 &     4 &    16 \\ 
  -94.34 &   512 &  0.00645 &    32 &  0.99447 &  0.99951 &     3 &    64 \\ 
\rowcolor{Gray}
 -134.43 &   256 &  0.00689 &   128 &  0.99961 &  0.99877 &     4 &    32 \\ 
 \hline 
\end{tabular}
\end{center}
\caption{
The parameter values and corresponding average returns for all 30 hyper-parameter configurations of PPO on UR-Reacher-2.
}
\end{table}

\newpage

\begin{figure}[h]
\includegraphics[scale=0.45]{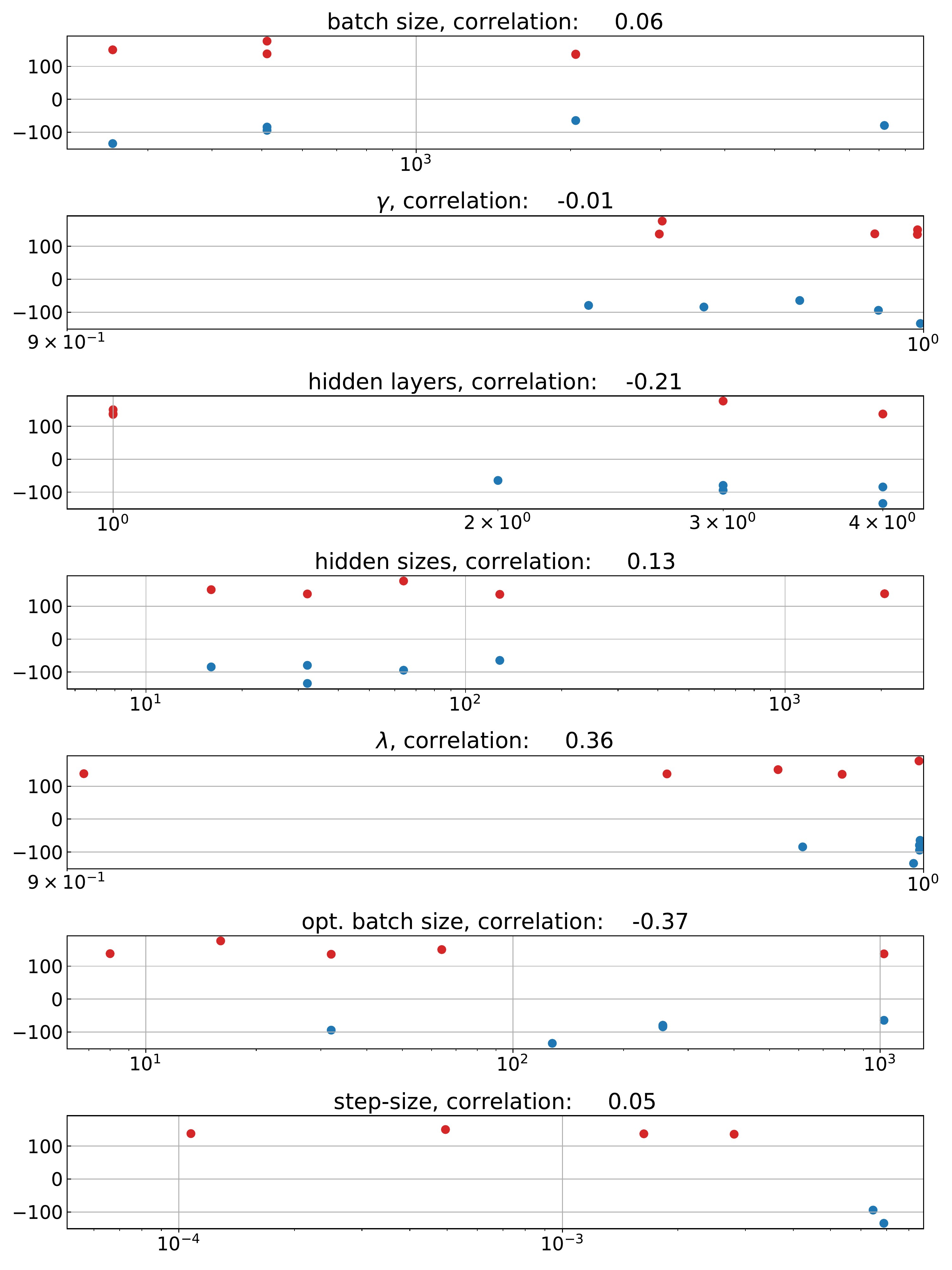}
\caption{
Average returns (y axis) vs. parameter values (x axis) for
best 5 and worst 5 hyper-parameter configurations of PPO on UR-Reacher-2.}
\label{fig:ur5_ppo_params}
\end{figure}

\newpage

\begin{table}[h]
\begin{center}
\begin{tabular}{ |c|c|c|c|c|c|c|c| } 
 \hline
 Average & batch & step-size & opt. & $\gamma$ & $\lambda$ & hidden & hidden \\
 Return & size & & batch size & & & layers & sizes \\
\hline
 -17.46 &   512 &  0.00680 &   128 &  0.99580 &  0.99183 &     3 &    32 \\ 
\rowcolor{Gray}
  -19.68 &  1024 &  0.00315 &   256 &  0.99304 &  0.99777 &     4 &    32 \\ 
  -20.34 &   512 &  0.00005 &    16 &  0.96366 &  0.99895 &     3 &    64 \\ 
\rowcolor{Gray}
  -20.45 &   256 &  0.00003 &    16 &  0.98987 &  0.97595 &     4 &   128 \\ 
  -20.57 &   512 &  0.00090 &   128 &  0.99195 &  0.99653 &     3 &     8 \\ 
\rowcolor{Gray}
  -20.91 &   512 &  0.00011 &     8 &  0.99161 &  0.90163 &     1 &  2048 \\ 
  -22.26 &  2048 &  0.00280 &    32 &  0.99864 &  0.98684 &     1 &   128 \\ 
\rowcolor{Gray}
  -22.56 &   256 &  0.00050 &    64 &  0.99867 &  0.97815 &     1 &    16 \\ 
  -23.45 &  2048 &  0.00163 &  1024 &  0.96331 &  0.96423 &     4 &    32 \\ 
\rowcolor{Gray}
  -24.66 &  4096 &  0.00036 &    64 &  0.99678 &  0.92655 &     4 &   128 \\ 
  -25.67 &  4096 &  0.00941 &  1024 &  0.98164 &  0.97644 &     4 &     8 \\ 
\rowcolor{Gray}
  -26.40 &  4096 &  0.00061 &   512 &  0.99812 &  0.99796 &     1 &     8 \\ 
  -26.43 &   512 &  0.00003 &    32 &  0.99524 &  0.92844 &     1 &   256 \\ 
\rowcolor{Gray}
  -26.68 &  4096 &  0.00080 &     8 &  0.99880 &  0.99558 &     3 &     8 \\ 
  -26.69 &  4096 &  0.00079 &    32 &  0.99699 &  0.99929 &     4 &    16 \\ 
\rowcolor{Gray}
  -27.14 &   256 &  0.00689 &   128 &  0.99924 &  0.99792 &     4 &    32 \\ 
  -27.66 &  8192 &  0.00226 &    32 &  0.96976 &  0.99472 &     4 &    64 \\ 
\rowcolor{Gray}
  -28.56 &  8192 &  0.00037 &    16 &  0.98820 &  0.98005 &     4 &    32 \\ 
  -28.56 &  4096 &  0.00267 &  4096 &  0.96363 &  0.99786 &     2 &    64 \\ 
\rowcolor{Gray}
  -32.47 &  8192 &  0.00006 &  2048 &  0.99610 &  0.98628 &     2 &   128 \\ 
  -32.65 &  8192 &  0.00362 &    32 &  0.96958 &  0.99622 &     2 &   128 \\ 
\rowcolor{Gray}
  -32.66 &  8192 &  0.00019 &    64 &  0.99512 &  0.99921 &     1 &    64 \\ 
  -33.00 &  2048 &  0.00002 &  1024 &  0.98102 &  0.99918 &     2 &   128 \\ 
\rowcolor{Gray}
  -34.32 &  2048 &  0.00002 &  2048 &  0.99175 &  0.98320 &     3 &   128 \\ 
  -34.34 &   512 &  0.00002 &   256 &  0.96874 &  0.98142 &     4 &    16 \\ 
\rowcolor{Gray}
  -34.46 &  8192 &  0.00087 &  2048 &  0.98611 &  0.96624 &     3 &     8 \\ 
  -34.56 &  8192 &  0.00987 &    32 &  0.99902 &  0.98921 &     2 &    16 \\ 
\rowcolor{Gray}
  -34.79 &  8192 &  0.00002 &   256 &  0.95497 &  0.99905 &     3 &    32 \\ 
  -37.00 &  1024 &  0.00039 &   256 &  0.99897 &  0.99924 &     3 &     8 \\ 
\rowcolor{Gray}
  -38.49 &   512 &  0.00645 &    32 &  0.99216 &  0.99907 &     3 &    64 \\ 
 \hline 
\end{tabular}
\end{center}
\caption{
The parameter values and corresponding average returns for all 30 hyper-parameter configurations of PPO on DXL-Reacher.
}
\end{table}

\newpage

\begin{figure}[h]
\includegraphics[scale=0.45]{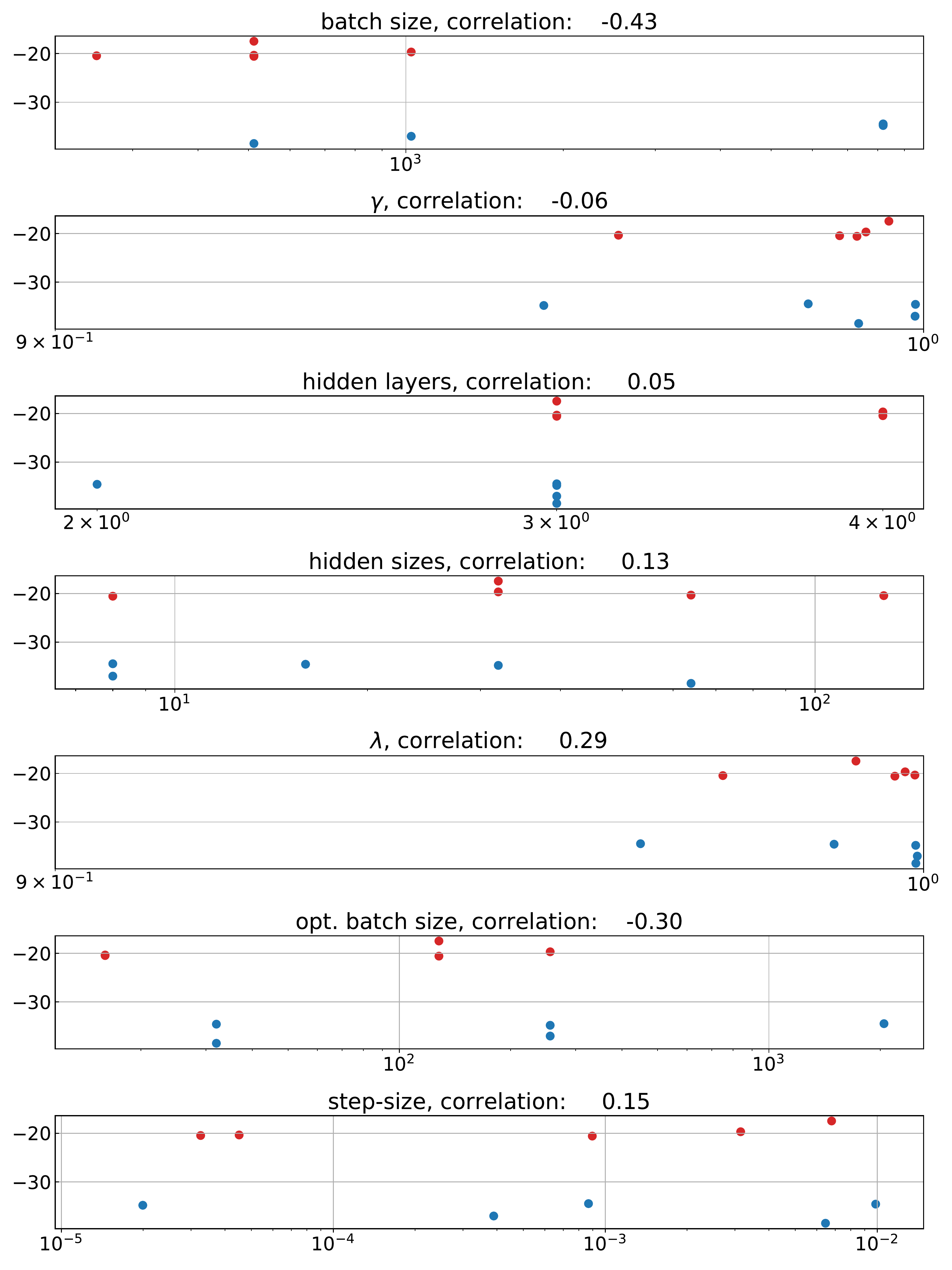}
\caption{
Average returns (y axis) vs. parameter values (x axis) for
best 5 and worst 5 hyper-parameter configurations of PPO on DXL-Reacher.}
\label{fig:ur5_ppo_params}
\end{figure}

\newpage

\begin{table}[h]
\begin{center}
\begin{tabular}{ |c|c|c|c|c|c|c|c| } 
 \hline
 Average & batch & step size & epochs & $\gamma$ & reward & hidden & hidden \\
 Return & size &  & & & scale & layers & sizes \\
\hline
186.99 &   256 &  0.00362 &     1 &  0.97415 & 17.24280 &     2 &   128 \\ 
\rowcolor{Gray}
  175.97 &   128 &  0.00267 &     2 &  0.96833 & 44.86500 &     2 &    64 \\ 
  160.74 &   256 &  0.00510 &     2 &  0.95963 &  8.34880 &     3 &    32 \\ 
\rowcolor{Gray}
  155.90 &   256 &  0.00226 &     1 &  0.97433 & 51.77960 &     4 &    64 \\ 
  152.24 &    64 &  0.00015 &     1 &  0.96801 &  9.17668 &     4 &    32 \\ 
\rowcolor{Gray}
  142.35 &   256 &  0.00987 &     1 &  0.99948 &  3.96065 &     2 &    16 \\ 
  141.76 &   128 &  0.00282 &     1 &  0.99813 &  7.31247 &     4 &    16 \\ 
\rowcolor{Gray}
  137.99 &    16 &  0.00034 &     4 &  0.97334 &  4.17074 &     4 &    16 \\ 
  128.37 &    16 &  0.00165 &     1 &  0.99402 & 15.26740 &     1 &  2048 \\ 
\rowcolor{Gray}
  127.41 &    32 &  0.00315 &     1 &  0.99516 &  7.33271 &     4 &    32 \\ 
  114.66 &   256 &  0.00005 &     4 &  0.98940 & 10.39050 &     3 &     8 \\ 
\rowcolor{Gray}
  105.15 &   256 &  0.00009 &     1 &  0.99750 &  3.29365 &     2 &   128 \\ 
   99.05 &    64 &  0.00002 &     2 &  0.99414 &  1.98432 &     3 &   128 \\ 
\rowcolor{Gray}
   95.70 &   128 &  0.00205 &     1 &  0.99935 & 18.56150 &     3 &     8 \\ 
   84.29 &    64 &  0.00028 &     4 &  0.99924 & 42.82680 &     1 &   128 \\ 
\rowcolor{Gray}
   81.99 &    64 &  0.00101 &     1 &  0.98488 &  1.33883 &     2 &   128 \\ 
   72.14 &   256 &  0.00019 &     2 &  0.99677 &  9.93572 &     1 &    64 \\ 
\rowcolor{Gray}
   67.54 &     8 &  0.00050 &     4 &  0.99926 &  4.92276 &     1 &    16 \\ 
   48.84 &   128 &  0.00222 &     2 &  0.98544 &  1.23942 &     4 &     8 \\ 
\rowcolor{Gray}
   44.17 &    16 &  0.00680 &     1 &  0.99728 &  3.10793 &     3 &    32 \\ 
   38.84 &    16 &  0.00645 &     4 &  0.99447 &  1.47516 &     3 &    64 \\ 
\rowcolor{Gray}
   27.85 &    16 &  0.00022 &     1 &  0.96836 &  1.31096 &     3 &    64 \\ 
  -20.55 &     8 &  0.00003 &     1 &  0.99260 & 19.36990 &     4 &   128 \\ 
\rowcolor{Gray}
  -26.10 &    16 &  0.00003 &     4 &  0.99686 &  3.89039 &     1 &   256 \\ 
  -26.97 &   256 &  0.00001 &     1 &  0.99119 &  1.36652 &     4 &    32 \\ 
\rowcolor{Gray}
  -42.94 &   128 &  0.00002 &     2 &  0.99799 & 23.75010 &     4 &   128 \\ 
  -85.84 &   128 &  0.00044 &     4 &  0.99891 & 31.71700 &     1 &     8 \\ 
\rowcolor{Gray}
  -87.05 &     8 &  0.00689 &     1 &  0.99961 & 16.78260 &     4 &    32 \\ 
  -87.38 &    16 &  0.00001 &     2 &  0.99430 &  1.11021 &     3 &     8 \\ 
\rowcolor{Gray}
  -87.52 &    32 &  0.00007 &     1 &  0.99945 &  8.72575 &     3 &     8 \\
 \hline 
\end{tabular}
\end{center}
\caption{
The parameter values and corresponding average returns for all 30 hyper-parameter configurations of Soft-Q on UR-Reacher-2.
}
\end{table}

\newpage

\begin{figure}[h]
\includegraphics[scale=0.45]{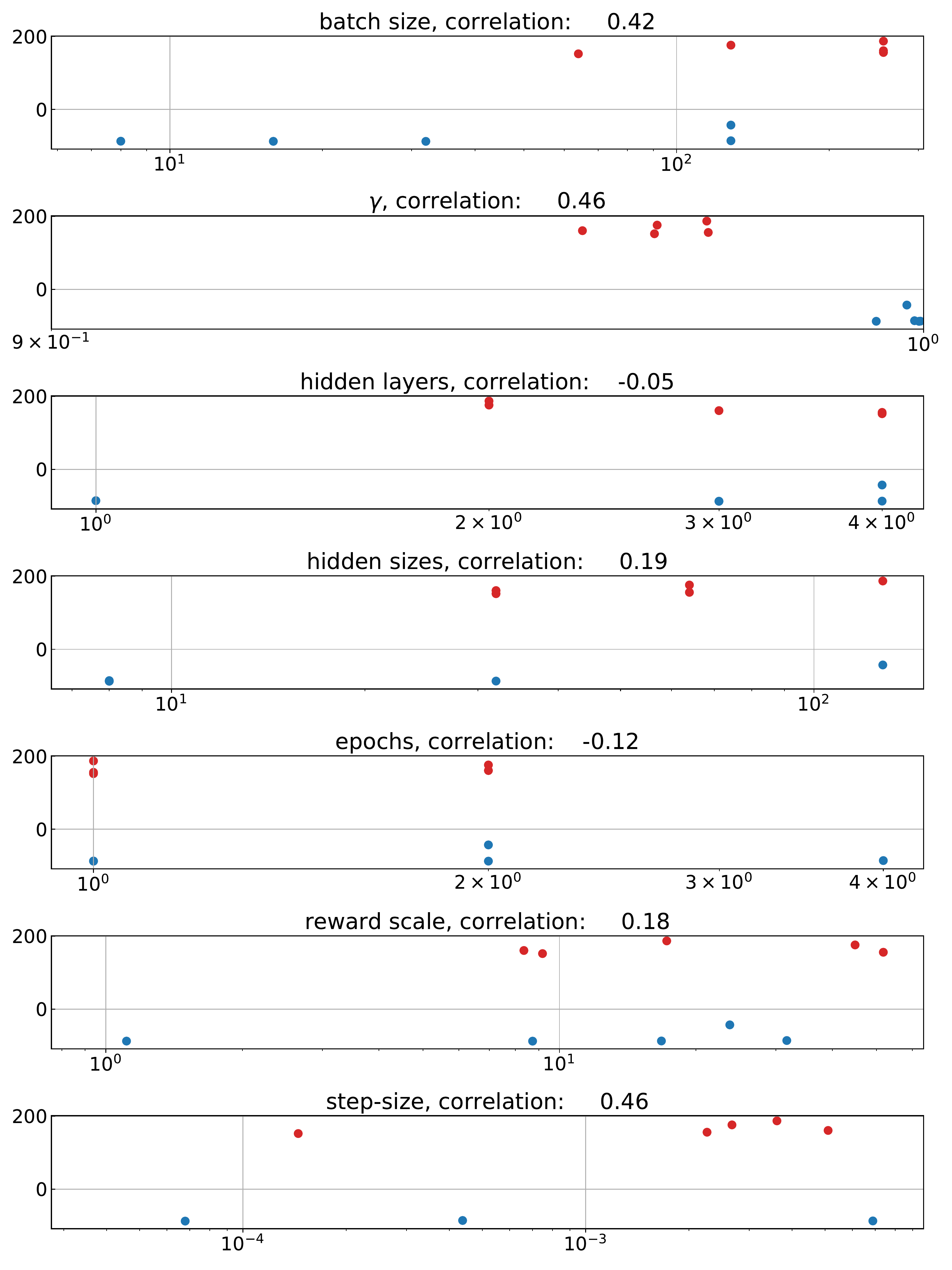}
\caption{
Average returns (y axis) vs. parameter values (x axis) for
best 5 and worst 5 hyper-parameter configurations of Soft-Q on UR-Reacher-2.}
\label{fig:ur5_ppo_params}
\end{figure}

\newpage

\begin{table}
\begin{center}
\begin{tabular}{ |c|c|c|c|c|c|c|c| } 
 \hline
 Average & batch & step size & epochs & $\gamma$ & reward & hidden & hidden \\
 Return & size &  & & & scale & layers & sizes \\
\hline
  -15.82 &   128 &  0.00002 &     2 &  0.99678 & 23.75010 &     4 &   128 \\ 
\rowcolor{Gray}
  -16.29 &   128 &  0.00267 &     2 &  0.96363 & 44.86500 &     2 &    64 \\ 
  -16.50 &    64 &  0.00028 &     4 &  0.99864 & 42.82680 &     1 &   128 \\ 
\rowcolor{Gray}
  -16.87 &   128 &  0.00205 &     1 &  0.99880 & 18.56150 &     3 &     8 \\ 
  -18.19 &    64 &  0.00015 &     1 &  0.96331 &  9.17668 &     4 &    32 \\ 
\rowcolor{Gray}
  -18.95 &   256 &  0.00226 &     1 &  0.96976 & 51.77960 &     4 &    64 \\ 
  -19.57 &   128 &  0.00282 &     1 &  0.99699 &  7.31247 &     4 &    16 \\ 
\rowcolor{Gray}
  -19.86 &    32 &  0.00315 &     1 &  0.99304 &  7.33271 &     4 &    32 \\ 
  -20.39 &   128 &  0.00044 &     4 &  0.99812 & 31.71700 &     1 &     8 \\ 
\rowcolor{Gray}
  -20.60 &     8 &  0.00689 &     1 &  0.99924 & 16.78260 &     4 &    32 \\ 
  -21.13 &    16 &  0.00165 &     1 &  0.99161 & 15.26740 &     1 &  2048 \\ 
\rowcolor{Gray}
  -21.43 &   256 &  0.00362 &     1 &  0.96958 & 17.24280 &     2 &   128 \\ 
  -22.39 &   256 &  0.00009 &     1 &  0.99610 &  3.29365 &     2 &   128 \\ 
\rowcolor{Gray}
  -22.60 &   256 &  0.00987 &     1 &  0.99902 &  3.96065 &     2 &    16 \\ 
  -22.89 &    16 &  0.00034 &     4 &  0.96874 &  4.17074 &     4 &    16 \\ 
\rowcolor{Gray}
  -23.32 &   256 &  0.00510 &     2 &  0.95497 &  8.34880 &     3 &    32 \\ 
  -23.76 &   256 &  0.00005 &     4 &  0.98611 & 10.39050 &     3 &     8 \\ 
\rowcolor{Gray}
  -24.50 &     8 &  0.00003 &     1 &  0.98987 & 19.36990 &     4 &   128 \\ 
  -25.37 &    16 &  0.00680 &     1 &  0.99580 &  3.10793 &     3 &    32 \\ 
\rowcolor{Gray}
  -25.44 &   256 &  0.00019 &     2 &  0.99512 &  9.93572 &     1 &    64 \\ 
  -26.35 &     8 &  0.00050 &     4 &  0.99867 &  4.92276 &     1 &    16 \\ 
\rowcolor{Gray}
  -28.30 &    64 &  0.00002 &     2 &  0.99175 &  1.98432 &     3 &   128 \\ 
  -28.45 &    16 &  0.00645 &     4 &  0.99216 &  1.47516 &     3 &    64 \\ 
\rowcolor{Gray}
  -29.49 &   128 &  0.00222 &     2 &  0.98164 &  1.23942 &     4 &     8 \\ 
  -30.06 &    64 &  0.00101 &     1 &  0.98102 &  1.33883 &     2 &   128 \\ 
\rowcolor{Gray}
  -31.12 &    16 &  0.00022 &     1 &  0.96366 &  1.31096 &     3 &    64 \\ 
  -32.15 &   256 &  0.00001 &     1 &  0.98820 &  1.36652 &     4 &    32 \\ 
\rowcolor{Gray}
  -33.83 &    32 &  0.00007 &     1 &  0.99897 &  8.72575 &     3 &     8 \\ 
  -33.89 &    16 &  0.00003 &     4 &  0.99524 &  3.89039 &     1 &   256 \\ 
\rowcolor{Gray}
  -34.35 &    16 &  0.00001 &     2 &  0.99195 &  1.11021 &     3 &     8 \\ 
 \hline 
\end{tabular}
\end{center}
\caption{
The parameter values and corresponding average returns for all 30 hyper-parameter configurations of Soft-Q on DXL-Reacher.
}
\end{table}

\newpage

\begin{figure}[h]
\includegraphics[scale=0.45]{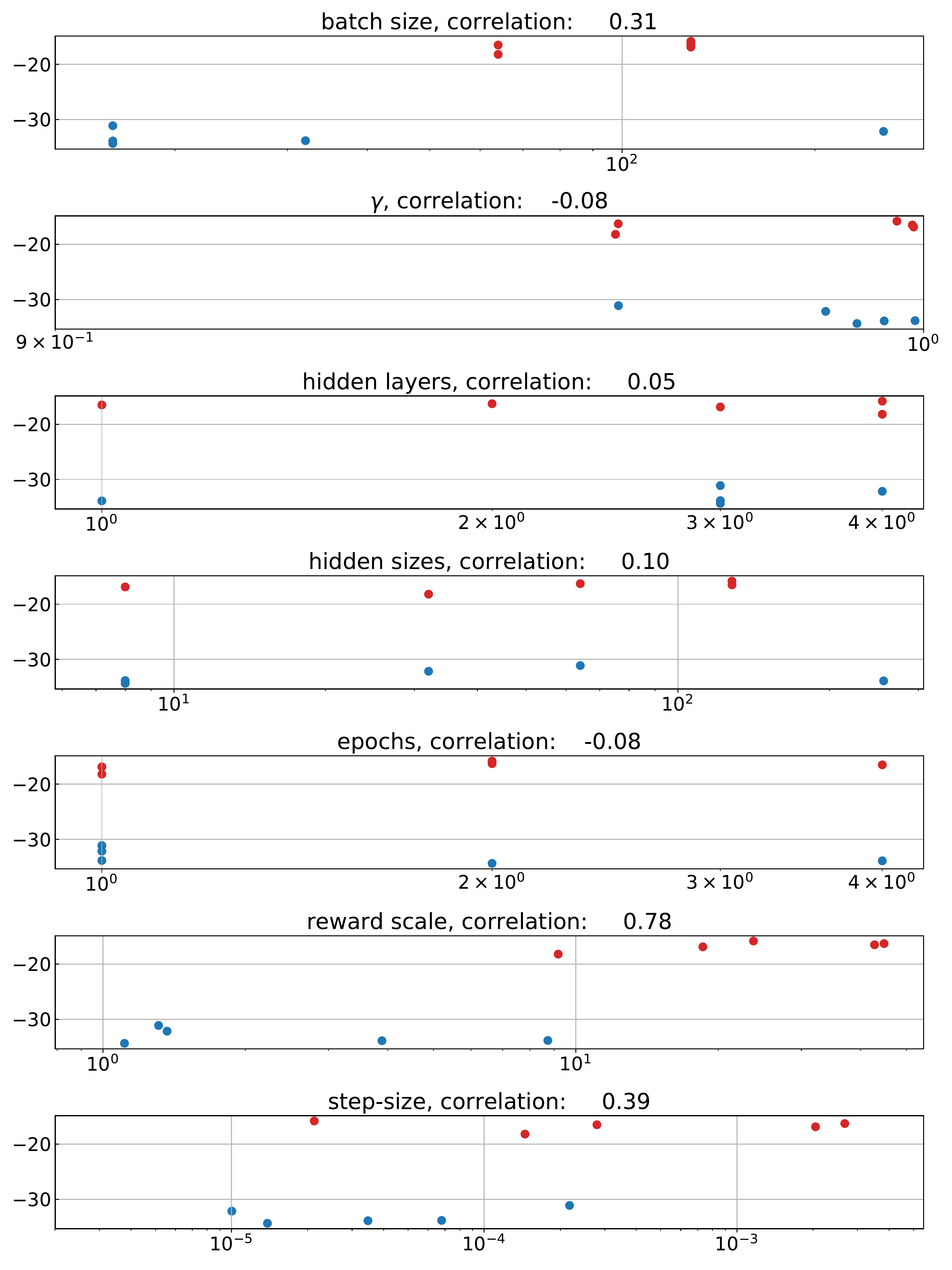}
\caption{
Average returns (y axis) vs. parameter values (x axis) for
best 5 and worst 5 hyper-parameter configurations of Soft-Q on DXL-Reacher.}
\label{fig:ur5_ppo_params}
\end{figure}

\newpage

\begin{table}[h]
\begin{center}
\begin{tabular}{ |c|c|c|c|c|c|c|c| } 
 \hline
 Average & batch & step size & exploration & $\gamma$ & reward & hidden & hidden \\
 Return & size &  & $\sigma$ & & scale & layers & sizes \\
\hline
   -5.16 &   128 &  0.00079 &  0.06797 &  0.97332 &  0.09400 &     2 &    64 \\ 
\rowcolor{Gray}
   -5.83 &    16 &  0.00077 &  0.10204 &  0.98430 &  2.02647 &     1 &     8 \\ 
  -22.24 &   128 &  0.00222 &  0.62454 &  0.98658 &  0.01536 &     1 &    16 \\ 
\rowcolor{Gray}
  -46.63 &   256 &  0.00019 &  0.12286 &  0.95695 &  0.98719 &     4 &    16 \\ 
  -59.76 &    16 &  0.00005 &  1.44343 &  0.99891 & 22.73270 &     1 &  2048 \\ 
\rowcolor{Gray}
  -77.93 &   256 &  0.00879 &  1.16057 &  0.99604 & 25.75380 &     1 &    32 \\ 
  -82.78 &    64 &  0.00002 &  0.65027 &  0.96275 &  4.68798 &     1 &  4096 \\ 
\rowcolor{Gray}
  -88.01 &   256 &  0.00002 &  1.87801 &  0.97475 & 40.72840 &     2 &    64 \\ 
  -90.53 &   128 &  0.00205 &  0.02196 &  0.99638 &  3.44530 &     4 &     8 \\ 
\rowcolor{Gray}
  -90.61 &     8 &  0.00004 &  0.30090 &  0.99967 &  0.04820 &     1 &    16 \\ 
  -91.54 &   256 &  0.00226 &  1.15201 &  0.99884 & 26.81130 &     4 &     8 \\ 
\rowcolor{Gray}
  -92.28 &   128 &  0.00016 &  1.44501 &  0.98836 & 17.21490 &     4 &    16 \\ 
  -92.46 &    16 &  0.00011 &  0.23885 &  0.99942 &  5.50558 &     1 &  4096 \\ 
\rowcolor{Gray}
  -92.92 &    16 &  0.00034 &  1.19977 &  0.99865 &  0.17395 &     4 &    32 \\ 
  -95.66 &    64 &  0.00163 &  1.46053 &  0.99909 &  0.02602 &     1 &  8192 \\ 
\rowcolor{Gray}
  -95.83 &     8 &  0.00065 &  0.02490 &  0.98394 &  2.36859 &     3 &   128 \\ 
  -97.56 &   256 &  0.00018 &  0.44355 &  0.99953 &  0.01111 &     1 &   128 \\ 
\rowcolor{Gray}
  -98.08 &   256 &  0.00001 &  0.01707 &  0.99033 & 98.53480 &     1 &   128 \\ 
  -98.48 &    32 &  0.00007 &  0.01814 &  0.98696 &  0.76139 &     4 &   128 \\ 
\rowcolor{Gray}
 -100.04 &    16 &  0.00095 &  0.11919 &  0.98391 &  0.05247 &     3 &    32 \\ 
 -101.92 &   256 &  0.00004 &  0.09315 &  0.99299 & 65.95210 &     4 &    64 \\ 
\rowcolor{Gray}
 -107.01 &   256 &  0.00002 &  0.36305 &  0.99880 &  1.24702 &     1 &    16 \\ 
 -108.30 &   128 &  0.00005 &  0.03798 &  0.95782 & 44.52130 &     3 &    32 \\ 
\rowcolor{Gray}
 -111.22 &     8 &  0.00161 &  0.01244 &  0.95337 &  0.02076 &     1 &  1024 \\ 
 -111.34 &   128 &  0.00002 &  0.07381 &  0.94064 &  5.64068 &     1 &    16 \\ 
\rowcolor{Gray}
 -117.14 &    64 &  0.00770 &  0.23378 &  0.99884 &  0.51910 &     2 &    16 \\ 
 -117.91 &    16 &  0.00251 &  0.21979 &  0.99915 &  0.02826 &     3 &    64 \\ 
\rowcolor{Gray}
 -119.55 &    32 &  0.00435 &  0.18999 &  0.98651 &  0.02679 &     1 &     8 \\ 
 -119.85 &    16 &  0.00001 &  0.22683 &  0.98117 &  0.01233 &     4 &    16 \\ 
\rowcolor{Gray}
 -123.91 &    64 &  0.00475 &  0.02562 &  0.99548 &  0.36455 &     4 &    64 \\ 
 \hline 
\end{tabular}
\end{center}
\caption{
The parameter values and corresponding average returns for all 30 hyper-parameter configurations of DDPG on UR-Reacher-2.
}
\end{table}

\newpage

\begin{figure}[h]
\includegraphics[scale=0.45]{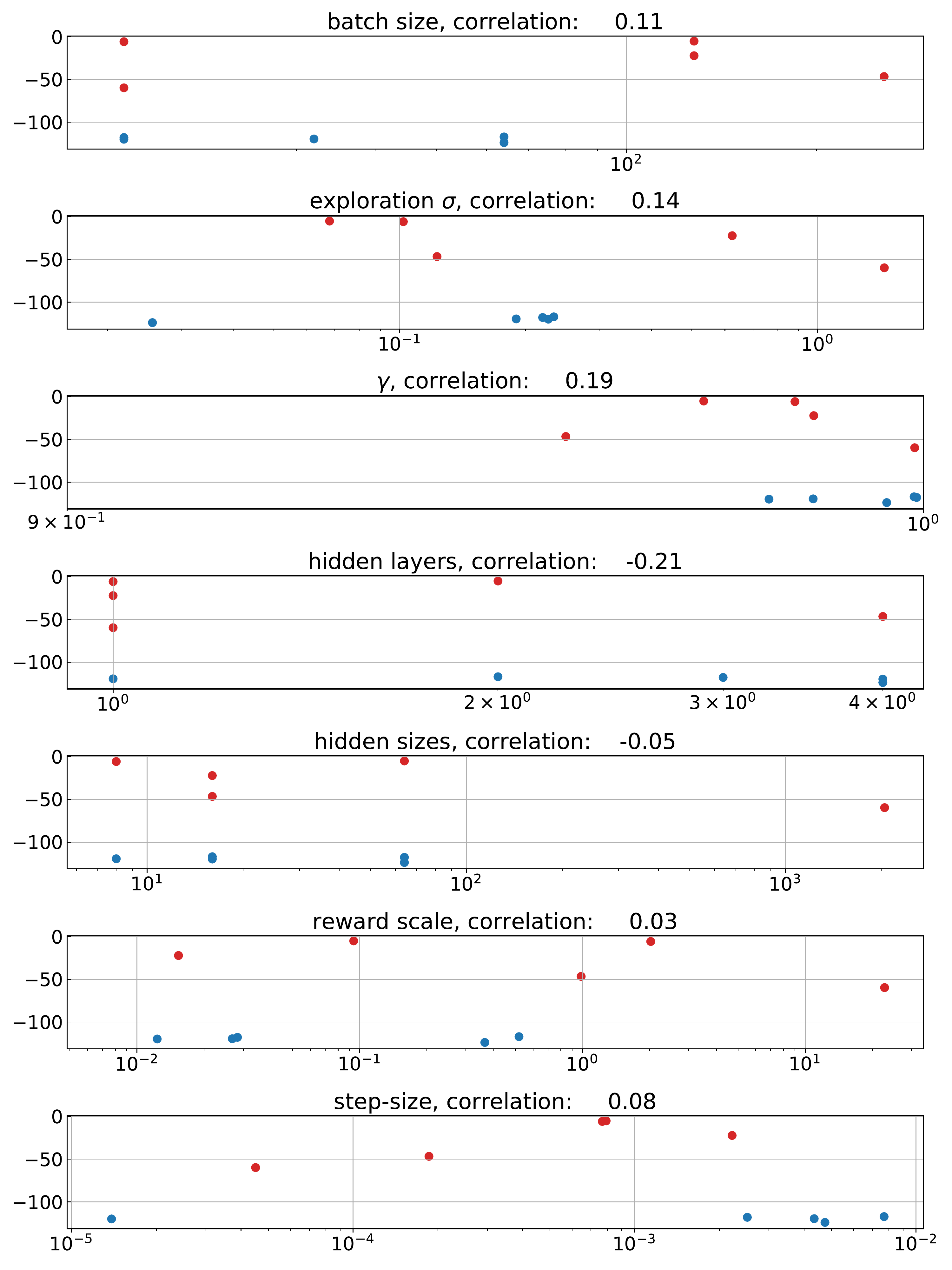}
\caption{
Average returns (y axis) vs. parameter values (x axis) for
best 5 and worst 5 hyper-parameter configurations of DDPG on UR-Reacher-2.}
\label{fig:ur5_ppo_params}
\end{figure}

\newpage 

\begin{table}
\begin{center}
\begin{tabular}{ |c|c|c|c|c|c|c|c| } 
 \hline
 Average & batch & step size & exploration & $\gamma$ & reward & hidden & hidden \\
 Return & size &  & $\sigma$ & & scale & layers & sizes \\
\hline
   -21.89 &    16 &  0.00077 &  0.10204 &  0.98038 &  2.02647 &     1 &     8 \\ 
\rowcolor{Gray}
  -23.52 &    32 &  0.00007 &  0.01814 &  0.98334 &  0.76139 &     4 &   128 \\ 
  -25.42 &   128 &  0.00205 &  0.02196 &  0.99460 &  3.44530 &     4 &     8 \\ 
\rowcolor{Gray}
  -27.41 &    64 &  0.00002 &  0.65027 &  0.95804 &  4.68798 &     1 &  4096 \\ 
  -28.71 &   256 &  0.00019 &  0.12286 &  0.95235 &  0.98719 &     4 &    16 \\ 
\rowcolor{Gray}
  -30.41 &   256 &  0.00002 &  0.36305 &  0.99796 &  1.24702 &     1 &    16 \\ 
  -30.80 &    16 &  0.00034 &  1.19977 &  0.99773 &  0.17395 &     4 &    32 \\ 
\rowcolor{Gray}
  -31.39 &   256 &  0.00879 &  1.16057 &  0.99416 & 25.75380 &     1 &    32 \\ 
  -32.06 &   128 &  0.00222 &  0.62454 &  0.98291 &  0.01536 &     1 &    16 \\ 
\rowcolor{Gray}
  -32.52 &    16 &  0.00005 &  1.44343 &  0.99812 & 22.73270 &     1 &  2048 \\ 
  -34.27 &   128 &  0.00016 &  1.44501 &  0.98492 & 17.21490 &     4 &    16 \\ 
\rowcolor{Gray}
  -34.90 &   256 &  0.00018 &  0.44355 &  0.99910 &  0.01111 &     1 &   128 \\ 
  -36.99 &    16 &  0.00011 &  0.23885 &  0.99893 &  5.50558 &     1 &  4096 \\ 
\rowcolor{Gray}
  -37.20 &   256 &  0.00001 &  0.01707 &  0.98719 & 98.53480 &     1 &   128 \\ 
  -37.61 &   256 &  0.00226 &  1.15201 &  0.99802 & 26.81130 &     4 &     8 \\ 
\rowcolor{Gray}
  -37.70 &    64 &  0.00163 &  1.46053 &  0.99839 &  0.02602 &     1 &  8192 \\ 
  -38.22 &    32 &  0.00435 &  0.18999 &  0.98283 &  0.02679 &     1 &     8 \\ 
\rowcolor{Gray}
  -38.23 &   128 &  0.00002 &  0.07381 &  0.93680 &  5.64068 &     1 &    16 \\ 
  -38.42 &   256 &  0.00002 &  1.87801 &  0.97020 & 40.72840 &     2 &    64 \\ 
\rowcolor{Gray}
  -38.67 &     8 &  0.00004 &  0.30090 &  0.99935 &  0.04820 &     1 &    16 \\ 
  -39.69 &    16 &  0.00001 &  0.22683 &  0.97697 &  0.01233 &     4 &    16 \\ 
\rowcolor{Gray}
  -40.63 &    64 &  0.00770 &  0.23378 &  0.99801 &  0.51910 &     2 &    16 \\ 
  -41.41 &   128 &  0.00079 &  0.06797 &  0.96872 &  0.09400 &     2 &    64 \\ 
\rowcolor{Gray}
  -42.16 &     8 &  0.00065 &  0.02490 &  0.97999 &  2.36859 &     3 &   128 \\ 
  -43.82 &    16 &  0.00095 &  0.11919 &  0.97995 &  0.05247 &     3 &    32 \\ 
\rowcolor{Gray}
  -44.08 &    16 &  0.00251 &  0.21979 &  0.99849 &  0.02826 &     3 &    64 \\ 
  -46.25 &    64 &  0.00475 &  0.02562 &  0.99344 &  0.36455 &     4 &    64 \\ 
\rowcolor{Gray}
  -48.74 &   256 &  0.00004 &  0.09315 &  0.99034 & 65.95210 &     4 &    64 \\ 
  -49.23 &   128 &  0.00005 &  0.03798 &  0.95320 & 44.52130 &     3 &    32 \\ 
\rowcolor{Gray}
  -49.98 &     8 &  0.00161 &  0.01244 &  0.94889 &  0.02076 &     1 &  1024 \\
 \hline 
\end{tabular}
\end{center}
\caption{
The parameter values and corresponding average returns for all 30 hyper-parameter configurations of DDPG on DXL-Reacher.
}
\end{table}

\newpage

\begin{figure}[h]
\includegraphics[scale=0.45]{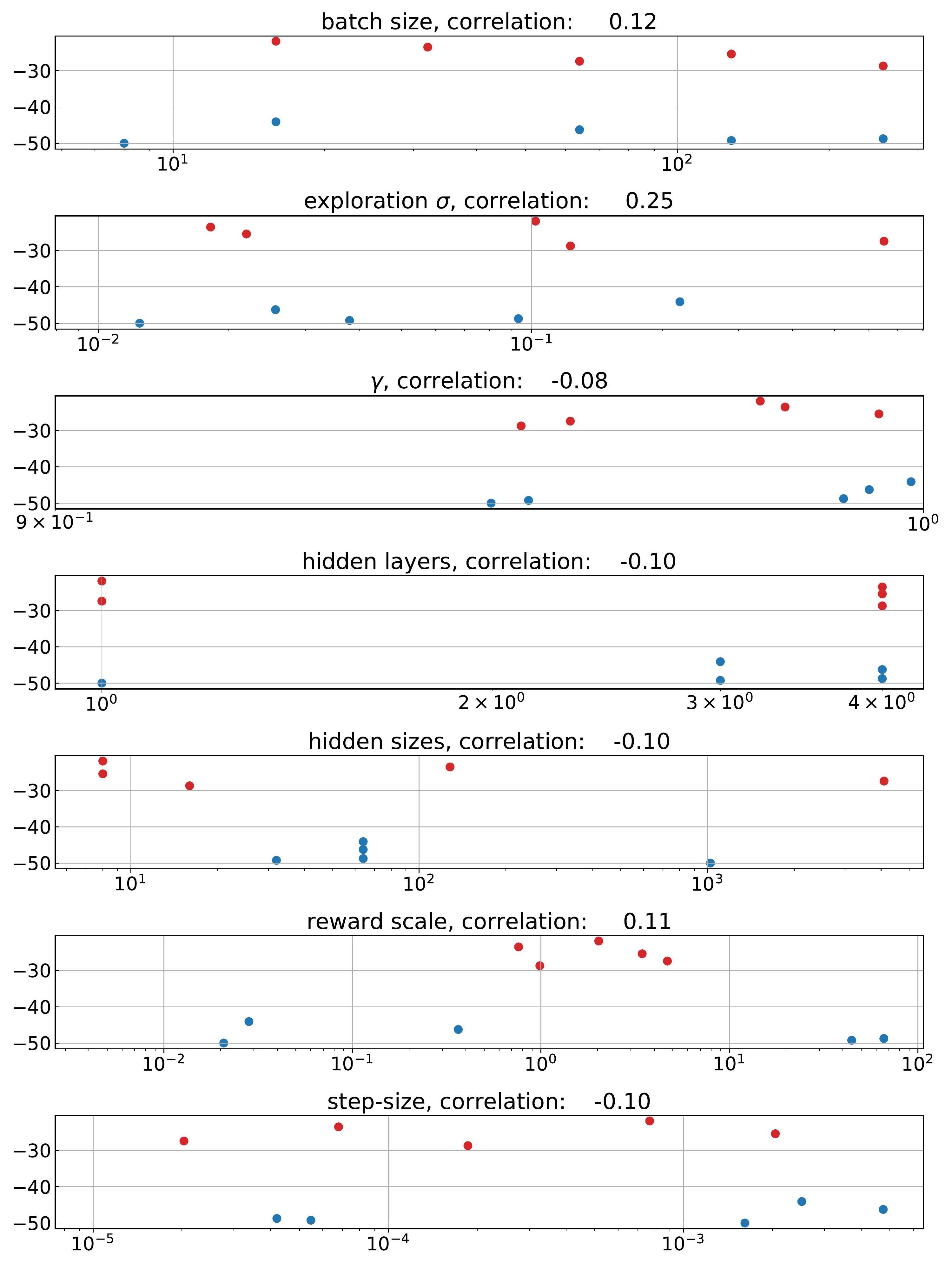}
\caption{
Average returns (y axis) vs. parameter values (x axis) for
best 5 and worst 5 hyper-parameter configurations of  DDPG on  DXL-Reacher.}
\label{fig:ur5_ppo_params}
\end{figure}

\end{document}